\documentclass{article}

\PassOptionsToPackage{table}{xcolor}
\usepackage{iclr2027_conference,times}
\iclrfinalcopy

\usepackage[utf8]{inputenc}
\usepackage[T1]{fontenc}
\usepackage{hyperref}
\usepackage{url}
\usepackage{booktabs}
\usepackage{amsfonts}
\usepackage{nicefrac}
\usepackage{microtype}
\usepackage[table]{xcolor}
\usepackage{xspace}

\usepackage{graphicx}
\usepackage{amsmath}
\usepackage{wrapfig}
\usepackage{algorithm}
\usepackage{algpseudocode}

\definecolor{cvprblue}{rgb}{0.21,0.49,0.74}
\hypersetup{colorlinks=true, linkcolor=cvprblue, citecolor=cvprblue, urlcolor=cvprblue}

\makeatletter
\DeclareRobustCommand\onedot{\futurelet\@let@token\@onedot}
\def\@onedot{\ifx\@let@token.\else.\null\fi\xspace}

\makeatother

\begin{document}

\title{BoxWorld: Let World Engines Control and Video Generators Render}
\author{%
    Zike Wu$^{1,3}$\quad
    Ahmed Hatem$^{1,2}$\quad
    Chintan Trivedi$^2$\quad
    Wallace Michel Pinto Lira$^2$\\
    \bfseries
    Xuebin Qin$^2$\quad
    Lele Wang$^1$\quad
    Renjie Liao$^{1,3,4}$\\
    $^1$University of British Columbia\quad
    $^2$Huawei Canada\\
    $^3$Vector Institute for AI\quad
    $^4$Canada CIFAR AI Chair
}

\maketitle
\lhead{}
\renewcommand{\headrulewidth}{0pt}
\begin{figure}[h]
    \centering
    \vspace{-10pt}
    \includegraphics[width=0.94\linewidth]{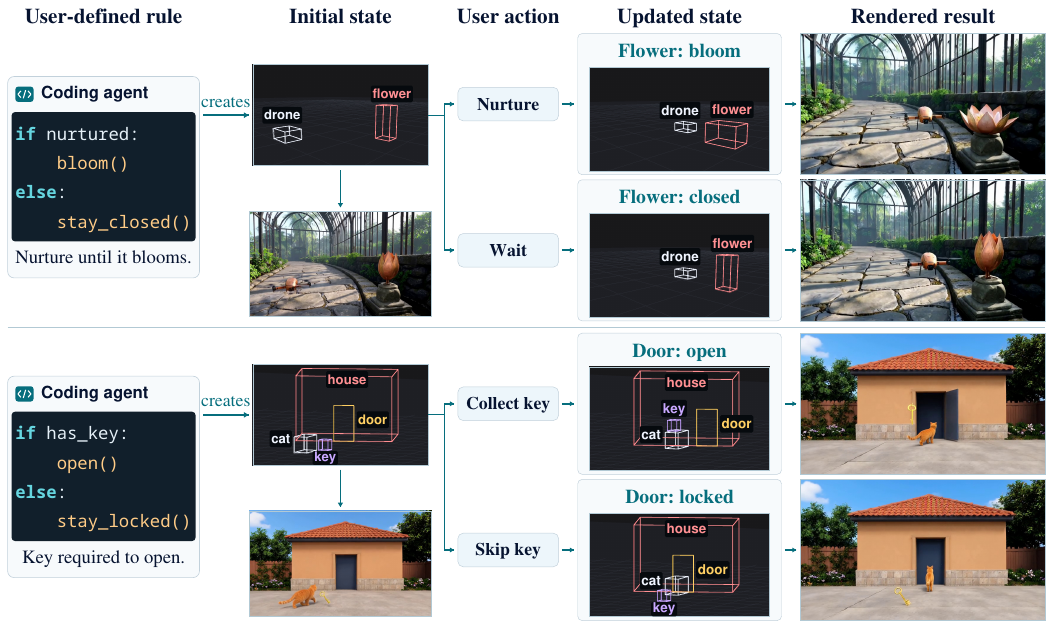}
    \vspace{-10pt}
    \caption{\textbf{BoxWorld separates interaction reasoning from generative rendering.} A coding agent creates a \textbf{Box World Engine} from user-defined rules. The engine updates object states from user actions, and the \textbf{Box-State Generative Renderer} renders them into video.}
    \label{fig:teaser}
\end{figure}
\suppressfloats[t]

\begin{abstract}
We tackle the challenge of building real-time interactive video worlds with user-defined behaviors. Such worlds must track object states, apply interaction rules, and render the resulting consequences as users act. Yet existing end-to-end video world models typically learn interaction dynamics within the generator, making it difficult to specify object-level rules and reason about their consequences. In this work, we introduce \texttt{BoxWorld}, which separates interaction reasoning from generative rendering. Given user-specified rules, a coding agent creates a \textbf{Box World Engine} that determines interaction outcomes and exposes the resulting object states through a lightweight interface of layout boxes, persistent object attributes, and dynamic status variables, leaving appearance and physical effects to generative rendering. A \textbf{Box-State Generative Renderer} then renders this evolving state stream into video autoregressively, enabling online control over object motion, state switches, and conditional events. Rendering this state stream is non-trivial: the renderer must respond to object-level state updates in real time while remaining coherent with the generated history. To this end, we introduce a three-stage training strategy that learns object grounding from open-domain videos and state transitions from gameplay videos, both annotated automatically, and finally performs rollout distillation on self-generated histories. We further propose PairedStateBench, a paired benchmark for fine-grained interaction control. BoxWorld substantially outperforms end-to-end world models and remains competitive across multiple public benchmarks.
\textbf{Project page:} \href{https://boxworld-web.github.io}{\texttt{boxworld-web.github.io}}.
\end{abstract}

\section{Introduction}

Building models that can understand, simulate, and interact with the physical world has long been an ambitious goal of artificial intelligence~\citep{worldmodel}.
Recent progress in video generation~\citep{wan,veo3,seeddance} has given this goal a concrete instantiation: \textit{interactive video worlds}, where users can act in a generated world and experience it unfolding over time in response to their actions~\citep{sora_world,genie,genie2,genie3}.
We are interested in real-time interactive video worlds with user-defined behaviors, where users also specify how the world responds to their actions.
For example, a user may specify that a flower blooms only when nurtured (Figure~\ref{fig:teaser}(top)), or that a door opens only if the cat has collected the key (Figure~\ref{fig:teaser}(bottom)).
Realizing such behaviors requires the world to track object states, apply interaction rules, and render the consequences as users act.

\begin{figure}[t]
    \centering
    \includegraphics[width=\linewidth]{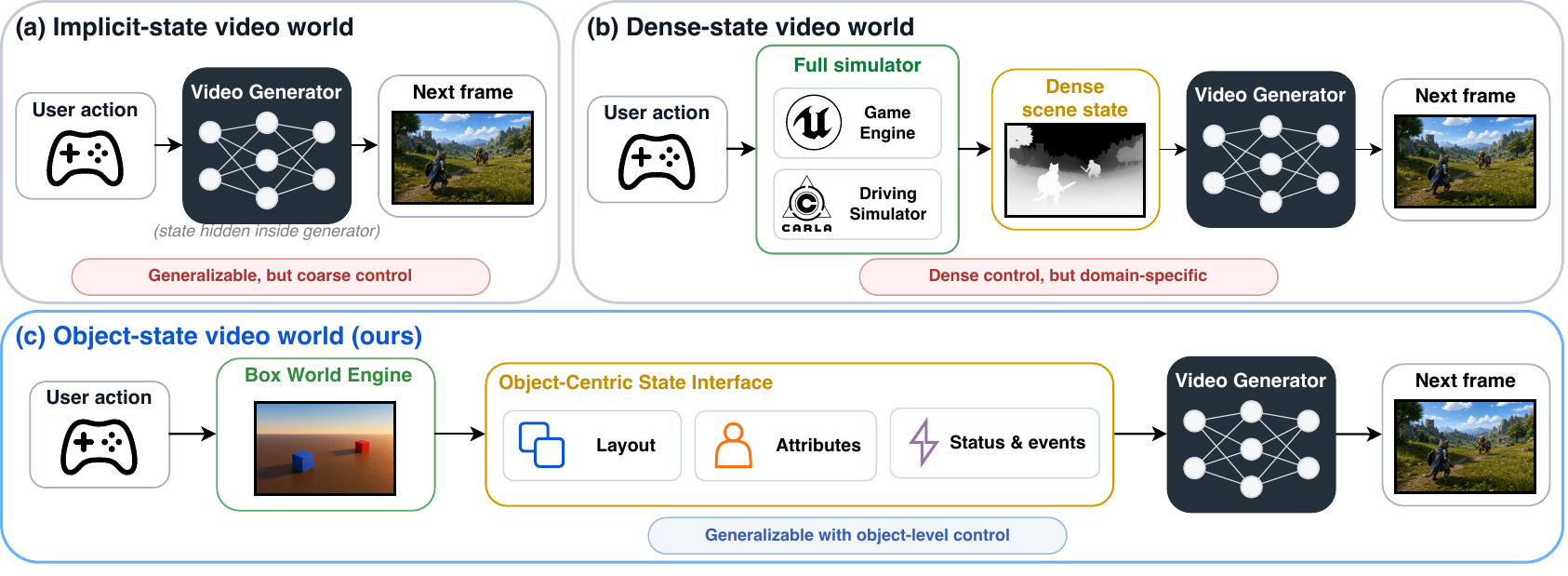}
    \vspace{-21pt}
    \caption{\textbf{Interfaces for interactive video worlds.}
(a) Implicit-state approaches benefit from broad visual generalization, but provide only coarse control.
(b) Dense-state approaches expose finer control through simulators, but such interfaces are often domain-specific and costly to construct and simulate.
(c) Our \texttt{BoxWorld} exposes an explicit-but-minimal \textbf{Object-Centric State Interface} for object-level control, while the video generator completes the remaining physical effects.}
    \label{fig:interfaces}
\end{figure}

A prevailing approach learns interactive video worlds end to end, mapping visual history and user actions directly to future frames with a video generator that generalizes across diverse scenes~\citep{genie,lingbot_world} (Figure~\ref{fig:interfaces}(a)).
However, the generator must then handle all three tasks implicitly, reasoning about whether rules are satisfied and what consequences should follow, which makes object-level rules and their consequences difficult to specify and control.
A complementary direction instead conditions video generation on explicit states, ranging from trajectories~\citep{trackdiffusion} and layouts~\citep{blobgen_vid} to physical priors~\citep{diffphy}, 3D caches~\citep{gen3c}, and full simulators~\citep{vid2sim} (Figure~\ref{fig:interfaces}(b)).
These states provide finer control, but grow heavier and more domain-specific as control becomes finer, making them expensive to construct and simulate, and difficult to obtain from videos at scale.

Taken together, these lines of work reveal a trade-off between generalization and fine-grained control in interactive video worlds.
This motivates separating interaction reasoning from generative rendering: interaction rules are applied explicitly outside the generator, while the generator retains its broad visual priors and focuses on faithfully rendering the resulting object states.
The two sides are connected by an explicit-but-minimal \textbf{Object-Centric State Interface}, which exposes only the variables needed for object-level interaction while remaining simple to specify, update, and apply across diverse scenes.
Crucially, these state variables should be identifiable from videos with off-the-shelf perception models, so that learning the renderer is not restricted to predefined simulators or scarce action-paired interaction data.

Following this design, we present \texttt{BoxWorld}, a framework for real-time interactive video worlds with user-defined behaviors.
Given user-specified rules, a coding agent creates a \textbf{Box World Engine}, which applies these rules to user actions, determines the resulting interaction outcomes, and updates object states.
The engine exposes these states through the \textbf{Object-Centric State Interface}, composed of layout boxes, persistent object attributes, and dynamic status variables, leaving texture, lighting, style, and physical effects such as smoke and splashing water to generative rendering.
A \textbf{Box-State Generative Renderer} then renders this evolving state stream into video autoregressively, enabling online control over object motion, state switches, and conditional events (Figure~\ref{fig:interfaces}(c)).

However, rendering this state stream is non-trivial: the renderer must respond to object-level state updates in real time while remaining coherent with the generated history.
First, states arrive online, so generation must be few-step and causal, whereas existing layout-to-video methods rely on multi-step bidirectional generation over the entire layout sequence~\citep{trackdiffusion,lvd,videotetris,blobgen_vid}.
Second, an update makes part of the generated history outdated, so the renderer must distinguish what to preserve from what to change.
Third, the renderer must synthesize the new state of the affected object, such as a door swinging open.
Existing video editing methods achieve similar localized changes, but rely on precise masks, costly inversion, or per-video optimization~\citep{vace,revideo,anyv2v,loraedit}, making them unsuitable for online rendering from coarse boxes.

To address these challenges, we build the \textbf{Box-State Generative Renderer} on an autoregressive video generator~\citep{longlive} and train it in three stages, each with a training scheme matched to its role (Sec.~\ref{sec:training_strategy}).
The first stage, \textit{Masked Layout Grounding}, learns frame-wise object grounding on open-domain videos~\citep{vidgen1m,4dnex} with Diffusion Forcing~\citep{chen2024diffusion}, whose independent per-frame noise levels make grounding hold regardless of the noise level of the temporal context.
The second stage, \textit{State Dynamics Adaptation}, learns state transitions on gameplay videos with teacher forcing, whose clean history depicts the preceding state, so that the renderer learns to change the affected object while continuing the rest of the scene.
The third stage, \textit{Rollout Distillation}, distills the renderer on its own rollouts with Self-Forcing-style distribution matching~\citep{self_forcing}, restoring few-step generation while adapting it to the self-generated histories encountered at inference.
The state annotations are extracted by off-the-shelf perception models~\citep{groundingdino,llava_next_video}, requiring no action-paired interaction logs (Sec.~\ref{sec:video_to_structure}).

We further introduce PairedStateBench, which evaluates user-defined behaviors through paired branches with distinct intended outcomes, measuring whether generated videos realize specified consequences and suppress unspecified ones.
\texttt{BoxWorld} substantially outperforms end-to-end world models on PairedStateBench, and remains competitive in short- and long-horizon video quality on VBench and VBench-Long~\citep{vbench}, and layout-to-video generation on YouTube-VIS~\citep{vis2021}, while running in real time.
Our key contributions are as follows:
\begin{itemize}
\item We propose \texttt{BoxWorld}, a framework for real-time interactive video worlds with user-defined behaviors, in which a \textbf{Box World Engine} created by a coding agent determines interaction outcomes and a \textbf{Box-State Generative Renderer} realizes them through the \textbf{Object-Centric State Interface}.

\item We introduce a three-stage training strategy with distinct roles: \textit{Masked Layout Grounding} and \textit{State Dynamics Adaptation} learn object grounding from open-domain videos and state transitions from gameplay videos, both annotated automatically, and \textit{Rollout Distillation} restores few-step generation on the renderer's own rollouts.

\item We propose PairedStateBench for fine-grained interaction control, on which \texttt{BoxWorld} substantially outperforms end-to-end world models while remaining competitive on public benchmarks for video quality and layout-to-video generation.
\end{itemize}

\section{Related Work}
\textbf{Interactive Video World Models.} Early controllable world models showed that playable dynamics can be learned directly from video~\citep{playable_video_generation} or prompt-conditioned game loops~\citep{promptable_game_models}. This line then scaled to tokenized or latent world models~\citep{ivideogpt,genie,navigation_world_models} and larger video corpora~\citep{genie2,genie3}, and more recently to interactive simulators built from powerful video generators with action~\citep{gamengen,gamegenx}, camera~\citep{vid2world}, and navigation control~\citep{gamefactory,dws,mineworld}. Recent extensions further target open environments, character-centered interaction, and longer-horizon memory~\citep{from_virtual_games_to_real_world_play,matrixgame2,lingbot_world,worldmem,aether}. In parallel, stronger backbones~\citep{ltx_video,wan} together with causalization and distillation~\citep{causvid,self_forcing,longlive} are beginning to enable real-time interactive systems~\citep{motionstream}. These works show strong visual interaction, but scene dynamics and intervention logic still largely reside in the generator~\citep{gamengen,vid2world,longlive}, making object-level rules and their consequences difficult to specify and control.

\textbf{State-Conditioned Generation.} Another line exposes explicit state and conditions image or video generation on it. Early grounded generation uses layouts and regions for object-level control~\citep{gligen,gvdiff,layoutdiffuse}, while ControlNet-style methods extend explicit conditioning to dense signals such as edges, depth, segmentation, and pose~\citep{controlnet,unicontrolnet,controlnet_xs}. Later methods extend this idea to planning~\citep{videodirectorgpt}, trajectories and tracklets~\citep{dragnuwa,trailblazer,trackdiffusion,flashmotion,zhang2026trajloom}, camera and motion control~\citep{motionctrl,i2vcontrol_camera,controlavideo,controlvideo,videocomposer}, and more compositional object guidance~\citep{image_conductor,blobgen_vid,dreamforge,trackgo,ati}. Recent physics-aware and simulator-grounded variants further improve consistency by injecting physical priors or planning~\citep{diffphy,physmaster,vlipp,physctrl}, or by conditioning generation on richer state such as simulators, depth, geometry, and 3D caches~\citep{vid2sim,generative_rendering,gen3c,worldwarp}. Together, these methods show that explicit conditioning improves controllability, but they typically generate a complete clip offline, using multi-step bidirectional denoising given the full condition sequence in advance~\citep{trackdiffusion,blobgen_vid,motionctrl}, or depend on heavier state and supervision~\citep{vid2sim,generative_rendering,gen3c,worldwarp}.
In contrast, BoxWorld renders in real time, consuming object states as they arrive with few-step causal generation, and realizes state changes during the rollout while remaining coherent with the generated history.

\section{Method}
\begin{figure}[t]
    \centering
    \includegraphics[width=\linewidth]{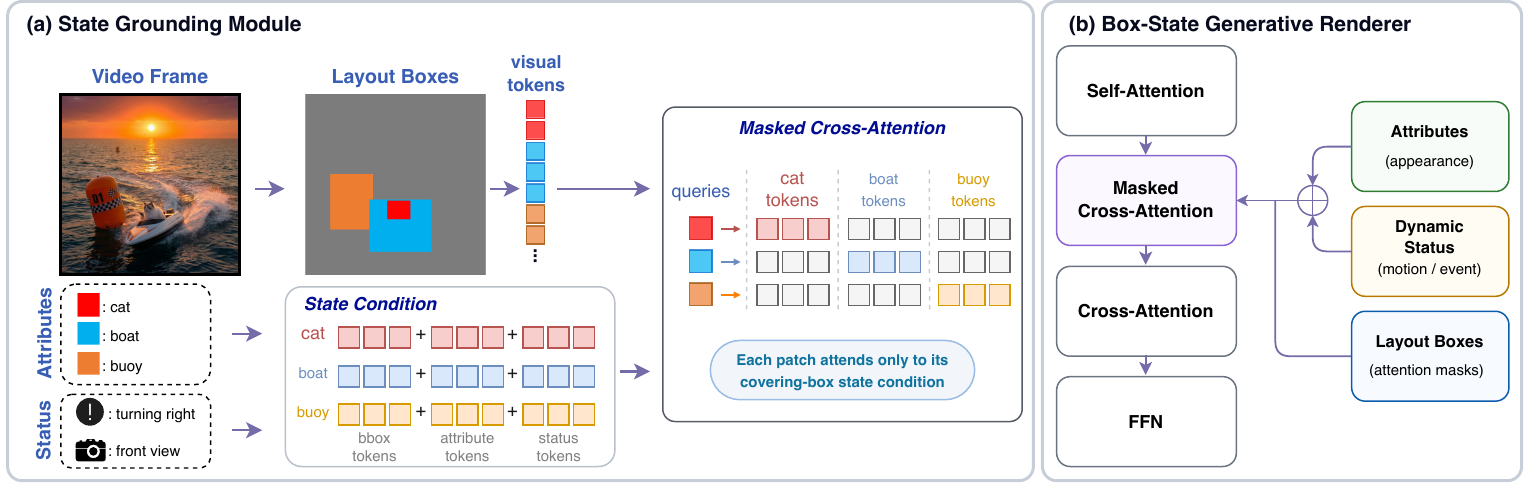}
    \vspace{-18pt}
    \caption{
    \textbf{Box-State Generative Renderer.}
    (a) The State Grounding Module encodes each object's box, attribute, and dynamic status into grounding tokens, and its masked cross-attention lets each visual token attend only to the objects whose boxes cover it.
    (b) The module is inserted between self-attention and cross-attention in each block of the renderer.
    }
    \label{fig:framework}
    \vspace{-12pt}
\end{figure}

\texttt{BoxWorld} assigns interaction reasoning to a \textbf{Box World Engine} and generative rendering to a \textbf{Box-State Generative Renderer}, which communicate through the \textbf{Object-Centric State Interface}. We describe the engine and interface in Sec.~\ref{sec:state_interface}, the renderer in Sec.~\ref{sec:state_renderer}, and how the renderer is trained in three stages (Sec.~\ref{sec:training_strategy}) on video-derived annotations (Sec.~\ref{sec:video_to_structure}).

\subsection{Box World Engine and Object-Centric State Interface}
\label{sec:state_interface}
At time step $t$, the Object-Centric State Interface describes the scene as $\mathcal{S}_t = \{(\mathbf{b}_{t,i}, \mathbf{c}_i, \mathbf{d}_{t,i})\}_{i=1}^{N_t}$ over $N_t$ active objects. For object $i$, the layout box $\mathbf{b}_{t,i}$ gives its 2D location in frame coordinates, the persistent attribute $\mathbf{c}_i$ describes its identity and appearance, and the dynamic status variable $\mathbf{d}_{t,i}$ describes its current status.

Given user-specified rules, a coding agent implements a Box World Engine from a shared engine template before interaction (Appendix~\ref{app:box_world_engine}). During interaction, the engine maintains its internal scene variables and updates them with each user action. It applies the rules to decide whether and when each object changes state, and projects the resulting object states into $\mathcal{S}_t$, which is published to a state buffer read asynchronously by the renderer.

Since $\mathcal{S}_t$ can be recovered from videos with off-the-shelf perception models (Sec.~\ref{sec:video_to_structure}), the renderer can be trained without engines or action-paired interaction logs, and at inference time it can be driven by any engine that publishes the same interface (Sec.~\ref{sec:engine_generalization}).

\subsection{Box-State Generative Renderer}
\label{sec:state_renderer}
\paragraph{Frame-wise state grounding.}
The Box-State Generative Renderer generates the latent video autoregressively, modeling $p_\theta(\mathbf{x}_t \mid \mathbf{x}_{<t}, \mathcal{S}_t)$, where $\mathbf{x}_t$ is the $t$-th latent frame, $\mathbf{x}_{<t}$ is the generated history, and $\mathcal{S}_t$ is the current state. Since $\mathcal{S}_t$ changes over time as objects move and change status, each frame should be grounded only with its own state. We therefore inject states frame by frame, which matches this constraint and avoids costly cross-frame attention. Specifically, we insert into each backbone block a State Grounding Module implemented as frame-wise masked cross-attention (Figure~\ref{fig:framework}). At frame $t$, the states of all objects are encoded into grounding tokens $\mathbf{E}_t \in \mathbb{R}^{(N_t L) \times D}$ with $L$ tokens per object, and their boxes define a mask $\mathbf{M}_t \in \{0,1\}^{HW \times (N_t L)}$ that lets each visual token attend only to the objects covering it. We reshape the flattened visual tokens $\mathbf{h} \in \mathbb{R}^{B \times (THW) \times D}$ into per-frame tokens $\tilde{\mathbf{h}} \in \mathbb{R}^{(BT) \times (HW) \times D}$ and apply masked cross-attention to each frame:
\begin{equation}
    \mathbf{Y}_t =
    \operatorname{Softmax}\left(
        \frac{
        (\tilde{\mathbf{h}}_t\mathbf{W}_Q)
        (\mathbf{E}_t\mathbf{W}_K)^\top
        }{\sqrt{D}}
        + \Phi(\mathbf{M}_t)
    \right)
    \mathbf{E}_t\mathbf{W}_V ,
\end{equation}
where $\Phi(\mathbf{M}_t)$ is $0$ for allowed entries and $-\infty$ otherwise. The per-frame outputs $\mathbf{Y}_t$ are reshaped back into the video-token sequence $\bar{\mathbf{Y}}$ and injected through a gated residual:
\begin{equation}
    \mathbf{h}' =
    \mathbf{h}
    + \lambda \tanh(\gamma)
    \bar{\mathbf{Y}},
\end{equation}
where $\lambda$ is a control scale and $\gamma$ is a learnable gate initialized to zero. Temporal coherence across frames is left to the causal self-attention of the backbone.

\paragraph{Asynchronous streaming rendering.}
At inference, the Box World Engine and the renderer run as asynchronous processes connected by the state buffer. The engine continuously receives user actions, updates its internal scene variables according to the rules, and publishes the resulting state $\mathcal{S}_t$ through the interface. Meanwhile, the renderer reads the latest state from the buffer whenever it generates new frames and renders them autoregressively from the generated history. A state update is therefore reflected in the next generated frames, while the engine keeps simulating the scene in parallel.

\subsection{Three-Stage Training}
\label{sec:training_strategy}
We initialize the renderer from LongLive~\citep{longlive}, a few-step autoregressive video generator distilled from Wan2.1~\citep{wan}, and train it in three stages: the first two learn object grounding and state transitions, respectively, and the third restores few-step generation by distilling on its own rollouts. In each stage, $\theta$ denotes the trainable parameters of that stage.

\paragraph{Stage 1: Masked Layout Grounding.}
This stage learns object grounding, binding each object to its layout box and persistent attribute, and trains on open-domain videos~\citep{vidgen1m,4dnex} whose diversity supports grounding across scenes. We freeze the backbone and train only the State Grounding Module. To bind every frame to its own state, we follow Diffusion Forcing~\citep{chen2024diffusion}, which assigns independent noise levels to the latent frames, so the module learns grounding regardless of the noise level of its temporal context. The objective is
\begin{equation}
    \mathcal{L}_{\mathrm{S1}}(\theta) =
    \mathbb{E}_{\mathbf{x}_0,t,\boldsymbol{\tau}}
    \left[
    w_{\mathrm{S1}}(\tau_t)
    \left\|
    \mathbf{v}_{\theta}
    \left(
    \mathbf{x}_{\tau_t,t},
    \tau_t,
    \mathbf{x}_{\boldsymbol{\tau}_{<t},<t},
    \mathcal{S}_{t}
    \right)
    - \mathbf{u}_{\tau_t,t}
    \right\|^2
    \right],
\end{equation}
where each frame is noised independently with levels $\boldsymbol{\tau}=(\tau_1,\dots,\tau_T)$, so the history $\mathbf{x}_{\boldsymbol{\tau}_{<t},<t}$ is itself noisy at varying levels, and $\mathbf{v}_{\theta}$ predicts the target velocity $\mathbf{u}_{\tau_t,t}$. Since boxes mainly constrain coarse layout, which is determined early in denoising, the weighting $w_{\mathrm{S1}}$~\citep{DiffSynth} emphasizes high noise levels, and for efficiency the module is active only at these levels during inference.

\paragraph{Stage 2: State Dynamics Adaptation.}
This stage learns state transitions, which open-domain videos rarely contain, so we train on gameplay videos. We freeze the Stage 1 renderer and train LoRA adapters~\citep{lora} in its linear layers, as LongLive~\citep{longlive} shows that LoRA adaptation of a distilled streaming generator approaches the quality of full fine-tuning. To learn how a status change alters the video, we use teacher forcing, in which the renderer denoises the current frame given the clean ground-truth prefix. When a status changes, the prefix depicts the preceding status and the target depicts the new one, so the adapters learn to change the affected object while continuing the rest of the scene. The objective is
\begin{equation}
    \mathcal{L}_{\mathrm{S2}}(\theta) =
    \mathbb{E}_{\mathbf{x}_0,t,\tau}
    \left[
    \left\|
    \mathbf{v}_{\theta}
    \left(
    \mathbf{x}_{\tau,t},
    \tau,
    \mathbf{x}_{0,<t},
    \mathcal{S}_{t}
    \right)
    - \mathbf{u}_{\tau,t}
    \right\|^2
    \right],
\end{equation}
where the history $\mathbf{x}_{0,<t}$ is the clean ground-truth prefix and $\mathbf{x}_{\tau,t}$ is the noisy current frame.

\paragraph{Stage 3: Rollout Distillation.}
The denoising objectives of the first two stages shift the renderer away from its distilled few-step distribution, and teacher forcing trains it on clean prefixes, whereas inference conditions on self-generated ones. This stage therefore restores few-step generation by distilling the renderer on its own rollouts. We freeze the Stage 2 renderer and train an additional LoRA adapter $\theta$ with Self-Forcing-style distribution matching distillation (DMD)~\citep{dmd,self_forcing}, using the Stage 2 renderer as the teacher and state sequences from both open-domain and gameplay annotations. The student generates few-step autoregressive rollouts $\hat{\mathbf{x}}$ and is updated with
\begin{equation}
    \nabla_{\theta}\mathcal{L}_{\mathrm{S3}} =
    \mathbb{E}_{\tau,\boldsymbol{\epsilon}}
    \left[
    w(\tau)
    \left(
    s_{\mathrm{fake}}(\hat{\mathbf{x}}_{\tau}, \tau, \mathcal{S}_{1:T})
    - s_{\mathrm{real}}(\hat{\mathbf{x}}_{\tau}, \tau, \mathcal{S}_{1:T})
    \right)^{\top}
    \frac{\partial \hat{\mathbf{x}}}{\partial \theta}
    \right],
\end{equation}
where $\hat{\mathbf{x}}_{\tau}$ is the rollout re-noised to level $\tau$, $s_{\mathrm{real}}$ and $s_{\mathrm{fake}}$ are the scores of the teacher and of a critic initialized from it, and $w(\tau)$ is the normalizer following DMD. Sec.~\ref{sec:ablation} evaluates its effect on long-horizon generation.

For efficient inference, we merge the LoRA adapters of LongLive and of Stages 2 and 3 into the backbone weights~\citep{lora}, so that the adapters add no runtime cost.

\subsection{Video-to-Structure Annotation}
\label{sec:video_to_structure}
We recover $\mathcal{S}_t$ from open-domain and gameplay videos with off-the-shelf perception models: LLaVA-NeXT-Video~\citep{llava_next_video} for clip descriptions, Grounding DINO~\citep{groundingdino} for initial layout boxes, SAM~2~\citep{sam2} for tracking, and object-level captioning for persistent attributes. Open-domain videos cover diverse objects and scenes but contain few status changes, whereas gameplay videos contain frequent actions and triggered events. For both sources, we crop each tracked object across frames and describe its current status and action with emphasis on status changes, so that the per-frame descriptions form $\mathbf{d}_{t,i}$ and expose status transitions. Appendix~\ref{app:annotation} provides details and examples.

\section{Experiments}
\subsection{Implementation Details}
\label{sec:experimental_setup}
We train Stage 1 on filtered open-domain videos from VideoGen-1M~\citep{vidgen1m} and 4DNeX~\citep{4dnex}, Stage 2 on 4,000 short gameplay clips collected online, and Stage 3 on state sequences from both sources. The three stages run for 20K, 10K, and 400 iterations, with rank-16 LoRA adapters in Stage 2 and a rank-64 adapter in Stage 3. At inference, the renderer uses four denoising steps at 832$\times$480, and the State Grounding Module is active only at noise levels above 0.8. The Box World Engines in our experiments are created by Kimi K2.5~\citep{kimik25} from user-specified rules (Appendix~\ref{app:implementation}).

\begin{table}[t]
    \centering
    \footnotesize
    \setlength{\tabcolsep}{4pt}
    \caption{\textbf{Results on PairedStateBench.} $^\dagger$Camera-control version, the only publicly released LingBot-v2 variant. Oracle variants additionally receive the interaction consequences as text.}
    \label{tab:authored_state}
    \begin{tabular}{lccccccc}
        \toprule
        Method & FPS$^{\uparrow}$ & Trigger$^{\uparrow}$ & Suppress$^{\uparrow}$ & Paired-CF$^{\uparrow}$ & State-Fid$^{\uparrow}$ & OnTime$^{\uparrow}$ & Task$^{\uparrow}$ \\
        \midrule
        \rowcolor{black!6}
        \multicolumn{8}{l}{\textit{Overall}} \\
        LingBot-v1 & 0.05 & 53.5 & 11.1 & 16.2 & 43.1 & 44.5 & 25.3 \\
        LingBot-v1 (Oracle) & 0.05 & 45.5 & 25.4 & 16.2 & 42.3 & 27.7 & 18.4 \\
        LingBot-v2$^\dagger$ & 13.3 & 21.3 & 63.5 & 10.8 & 46.1 & 14.3 & 11.5 \\
        LingBot-v2$^\dagger$ (Oracle) & 13.3 & 19.0 & 69.8 & 16.2 & 57.1 & 17.5 & 11.5 \\
        LongLive & \textbf{20.7} & 10.7 & 50.8 & 13.5 & 36.2 & 5.7 & 8.0 \\
        BoxWorld (Ours) & \textbf{20.7} & \textbf{92.2} & \textbf{90.5} & \textbf{97.3} & \textbf{91.6} & \textbf{91.2} & \textbf{83.9} \\
        \midrule
        \rowcolor{black!6}
        \multicolumn{8}{l}{\textit{Easy}} \\
        LingBot-v1 & 0.05 & 75.0 & -- & 60.0 & 38.3 & 75.0 & 47.8 \\
        LingBot-v1 (Oracle) & 0.05 & 50.0 & -- & 30.0 & 5.3 & 50.0 & 26.1 \\
        LingBot-v2$^\dagger$ & 13.3 & 25.0 & -- & 20.0 & 6.9 & 25.0 & 30.4 \\
        LingBot-v2$^\dagger$ (Oracle) & 13.3 & 50.0 & -- & 40.0 & 50.0 & 50.0 & 39.1 \\
        LongLive & \textbf{20.7} & 0.0 & -- & 30.0 & 0.0 & 0.0 & 26.1 \\
        BoxWorld (Ours) & \textbf{20.7} & \textbf{100.0} & -- & \textbf{100.0} & \textbf{98.4} & \textbf{100.0} & \textbf{82.6} \\
        \midrule
        \rowcolor{black!6}
        \multicolumn{8}{l}{\textit{Medium}} \\
        LingBot-v1 & 0.05 & 55.4 & 20.0 & 0.0 & 51.9 & 43.1 & 32.1 \\
        LingBot-v1 (Oracle) & 0.05 & 50.4 & 20.0 & 0.0 & 53.4 & 25.4 & 25.0 \\
        LingBot-v2$^\dagger$ & 13.3 & 26.1 & \textbf{100.0} & 0.0 & 53.6 & 10.9 & 7.1 \\
        LingBot-v2$^\dagger$ (Oracle) & 13.3 & 15.2 & \textbf{100.0} & 0.0 & 59.9 & 12.0 & 3.6 \\
        LongLive & \textbf{20.7} & 13.0 & \textbf{100.0} & 0.0 & 52.6 & 6.5 & 3.6 \\
        BoxWorld (Ours) & \textbf{20.7} & \textbf{85.9} & \textbf{100.0} & \textbf{100.0} & \textbf{90.2} & \textbf{85.9} & \textbf{89.3} \\
        \midrule
        \rowcolor{black!6}
        \multicolumn{8}{l}{\textit{Hard}} \\
        LingBot-v1 & 0.05 & 47.8 & 8.3 & 0.0 & 36.7 & 40.6 & 5.6 \\
        LingBot-v1 (Oracle) & 0.05 & 39.9 & 27.1 & 15.8 & 37.7 & 26.1 & 8.3 \\
        LingBot-v2$^\dagger$ & 13.3 & 15.9 & 52.1 & 10.5 & 44.6 & 15.9 & 2.8 \\
        LingBot-v2$^\dagger$ (Oracle) & 13.3 & 17.4 & 60.4 & 10.5 & 55.7 & 17.4 & 0.0 \\
        LongLive & \textbf{20.7} & 10.1 & 35.4 & 10.5 & 27.5 & 5.8 & 0.0 \\
        BoxWorld (Ours) & \textbf{20.7} & \textbf{97.1} & \textbf{87.5} & \textbf{94.7} & \textbf{91.9} & \textbf{94.9} & \textbf{80.6} \\
        \bottomrule
    \end{tabular}
\vspace{-10pt}
\end{table}

\subsection{Fine-Grained Interaction Control}
\label{sec:engine_generalization}

We evaluate fine-grained interaction control on PairedStateBench, whose cases range from directional motion and object handling to credential-dependent doors, charging, and multi-stage interactions. Each case specifies the intended state evolution, event conditions, and temporal windows, and paired cases share the initial scene but differ in the intended outcome. We compare against the end-to-end world models LingBot-v1~\citep{lingbot_world} and LingBot-v2~\citep{lingbot_world_v2}, each with an oracle variant that additionally receives the interaction consequences as text, and against LongLive~\citep{longlive}, the backbone of our renderer. The metrics measure whether required events appear (Trigger), forbidden consequences remain absent (Suppress), both branches of a pair succeed (Paired-CF), state windows and event onsets match the specification (State-Fid, OnTime), and complete multi-stage tasks succeed (Task). Appendix~\ref{app:authored_protocol} provides the full benchmark details.

\begin{figure}[p]
    \centering
    \includegraphics[width=\linewidth]{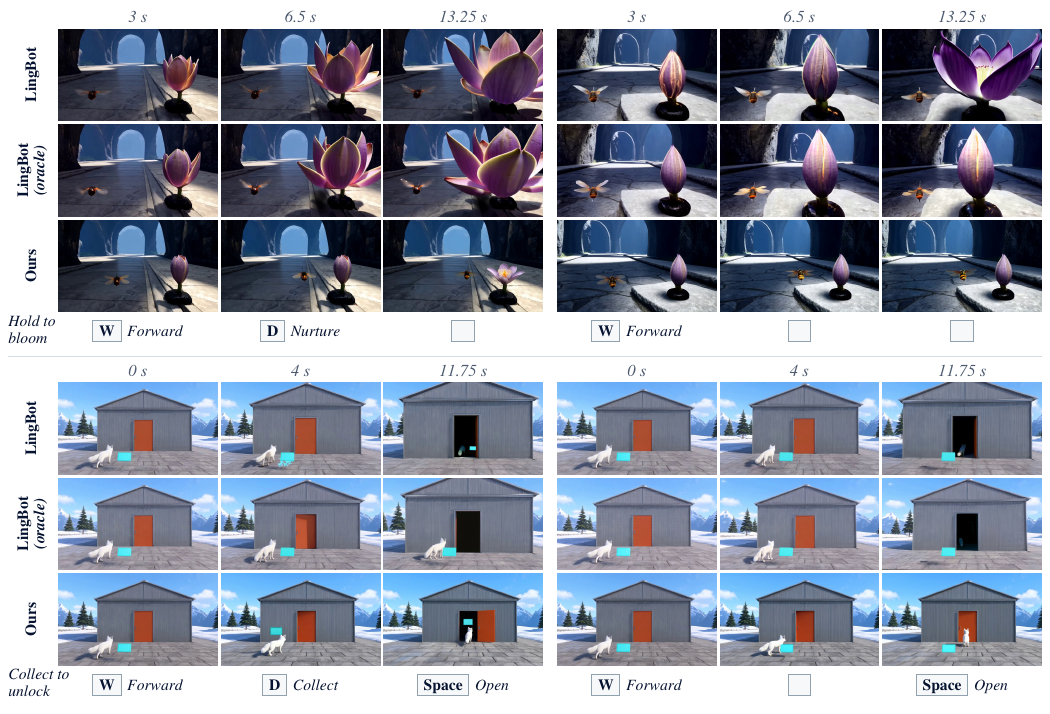}
    \vspace{-20pt}
    \caption{\textbf{Qualitative comparison on PairedStateBench.} Each case compares two branches from the same scene that differ in whether the precondition is met: nurturing the flower before it blooms (top) and collecting the key before opening the door (bottom). BoxWorld renders the event only once the precondition is met, whereas LingBot-v1 renders it without the precondition, and its oracle variant still opens the door without the key and blooms the flower before nurturing. More cases are in Appendix~\ref{app:authored_protocol}. Please refer to the \href{https://boxworld-web.github.io}{project page} for videos.}
    \label{fig:paired_comparison}
    \vspace{-15pt}
\end{figure}

\textbf{Quantitative Results.} As shown in Table~\ref{tab:authored_state}, BoxWorld realizes both the occurrence and the suppression of specified consequences, reaching 92.2\% Trigger and 90.5\% Suppress, and satisfies both branches in \textbf{97.3\%} of pairs, whereas no baseline exceeds \textbf{16.2\%} Paired-CF. The end-to-end models tend to produce the likely event regardless of the branch condition (Figure~\ref{fig:paired_comparison}), and supplying the resolved outcome as text does not close this gap: both oracle variants remain at 16.2\% Paired-CF. LongLive reaches only 13.5\%, indicating that the specified outcomes are realized through state grounding. BoxWorld also completes \textbf{83.9\%} of the full multi-stage tasks, compared with at most \textbf{25.3\%} for the baselines, and the Task gap widens with difficulty: on hard cases, BoxWorld retains 94.7\% Paired-CF and \textbf{80.6\%} Task, whereas the baselines reach at most 15.8\% and \textbf{8.3\%}, respectively. These results show that fine-grained interaction consequences, which end-to-end models struggle to infer from visual history, are straightforward for an explicit engine to determine, and that the Box-State Generative Renderer faithfully realizes them once they are expressed as object-level states.

\begin{figure}[p]
    \centering
    \includegraphics[width=\linewidth]{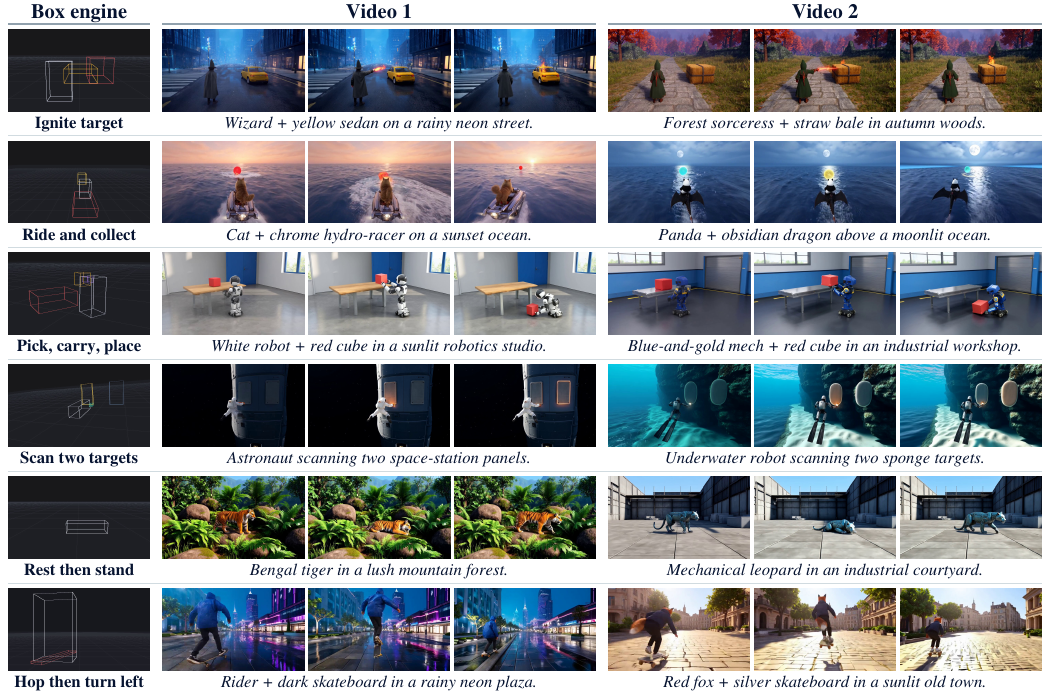}
    \vspace{-20pt}
    \caption{\textbf{Qualitative results of BoxWorld.} Each row shows a Box World Engine (left) and two videos rendered from the same state stream with different object attributes and scene descriptions. Please refer to the \href{https://boxworld-web.github.io}{project page} for videos.}
    \label{fig:qualitative}
    \vspace{-18pt}
\end{figure}

\textbf{Qualitative Results.} Figure~\ref{fig:paired_comparison} compares branches that differ in whether the precondition is met, such as nurturing the flower or collecting the key. BoxWorld renders the event only once the precondition holds, whereas LingBot-v1 renders it before the precondition is met, even with oracle text. Figure~\ref{fig:qualitative} further shows rollouts from six engines, each rendered in two visual worlds. The same interaction, such as picking up, carrying, and placing an object, is realized consistently across scenes and styles, indicating that one engine can drive diverse visual worlds through the Object-Centric State Interface.

\subsection{Public Benchmarks}
\label{sec:long_horizon}
\textbf{Video Quality.} We evaluate how much of the pretrained visual prior the renderer retains on VBench and 30-second VBench-Long~\citep{vbench}. As shown in Table~\ref{tab:video_quality} (Appendix~\ref{app:additional_video_quality}), BoxWorld achieves video quality comparable to LongLive at the same 20.7 FPS while improving the Semantic score from 76.47 to 79.54 on VBench and from 75.82 to 77.67 on VBench-Long, suggesting that object-level states strengthen compositional control.

\textbf{Layout-to-Video Generation.} Following BlobGEN-Vid~\citep{blobgen_vid}, we test on YouTube-VIS 2021, where ground-truth tracks provide layout boxes and per-track descriptions provide object attributes. As shown in Table~\ref{tab:youtubevis_grounding} (Appendix~\ref{app:ytvis_protocol}), BoxWorld is the only real-time method, yet it achieves the lowest FVD (62), the highest regional text and image alignment (rCLIP$_t$ 0.291, rCLIP$_i$ 0.837), and the highest regional cross-frame consistency (rCFC 0.965) among all compared methods. Its mIOU of 0.572 is second only to BlobGEN-Vid (0.598), whose blob representation offers finer spatial support than boxes.

\subsection{Ablation Studies}
\label{sec:ablation}
\begin{wraptable}{r}{0.47\linewidth}
    \vspace{-22pt}
    \centering
    \footnotesize
    \setlength{\tabcolsep}{4pt}
    \caption{\textbf{Ablations of training stages and the state interface.} Full tables are in Appendix~\ref{app:additional_experiments}.}
    \label{tab:ablation}
    \begin{tabular}{lcc}
        \toprule
        \rowcolor{black!6}
        \multicolumn{3}{l}{\textit{VBench}} \\
        Variant & Total$^{\uparrow}$ & Long Total$^{\uparrow}$ \\
        \midrule
        Stage 1 & 83.30 & 81.72 \\
        Stage 2 & 83.57 & 81.82 \\
        Stage 3 & \textbf{83.76} & \textbf{83.00} \\
        \midrule
        \rowcolor{black!6}
        \multicolumn{3}{l}{\textit{YouTube-VIS}} \\
        Variant & mIOU$^{\uparrow}$ & FVD$^{\downarrow}$ \\
        \midrule
        Stage 3 & \textbf{0.572} & \textbf{62} \\
        \quad w/o spatial mask & 0.291 & 79 \\
        \quad w/o object attributes & 0.323 & 68 \\
        \midrule
        \rowcolor{black!6}
        \multicolumn{3}{l}{\textit{PairedStateBench}} \\
        Variant & Suppress$^{\uparrow}$ & Task$^{\uparrow}$ \\
        \midrule
        Stage 1 & 60.3 & 49.4 \\
        Stage 2 & 85.7 & 64.4 \\
        Stage 3 & \textbf{90.5} & \textbf{83.9} \\
        \midrule
        Variant & Trigger$^{\uparrow}$ & OnTime$^{\uparrow}$ \\
        \midrule
        Stage 3 & \textbf{92.2} & \textbf{91.2} \\
        \quad w/o dynamic status & 42.5 & 16.1 \\
        \bottomrule
    \end{tabular}
    \vspace{-10pt}
\end{wraptable}

\textbf{Training Stages.} As shown in Table~\ref{tab:ablation}, the Total score on VBench and VBench-Long increases with each stage, and Stage 3 brings the largest gain on VBench-Long, from 81.82 to 83.00, showing that rollout distillation benefits long-horizon generation. On PairedStateBench, Stage 2 raises Suppress from 60.3\% to 85.7\% and Task from 49.4\% to 64.4\%, indicating that gameplay adaptation teaches the renderer to change an object only when its status changes. Stage 3 further raises Paired-CF from 67.6\% to 97.3\% and Task to 83.9\%, indicating that rollout distillation keeps state changes faithful over self-generated histories. Appendix~\ref{app:additional_ablations} reports all metrics and further compares Stage 2 adapter variants.

\textbf{State Interface.} On YouTube-VIS, removing the spatial mask lets every visual token attend to all objects and reduces mIOU from 0.572 to 0.291, which supports routing each object's state to its own region, and replacing all object attributes with the generic word ``object'' reduces mIOU to 0.323, showing that attributes identify objects beyond their box geometry. On PairedStateBench, replacing dynamic status with static descriptions lowers Trigger from 92.2\% to 42.5\% and OnTime from 91.2\% to 16.1\%, since dynamic status specifies whether and when each object changes.

\section{Conclusion}
We presented \texttt{BoxWorld}, which separates interaction reasoning from generative rendering for real-time interactive video worlds. A Box World Engine, created by a coding agent from user-specified rules, determines interaction outcomes and exposes them through the Object-Centric State Interface of layout boxes, persistent attributes, and dynamic status variables. A Box-State Generative Renderer, trained in three stages on automatically annotated videos, then renders the resulting state stream into video in real time. On PairedStateBench, BoxWorld realizes fine-grained interaction consequences that end-to-end world models struggle to produce, while retaining video quality comparable to its backbone. These results suggest a practical path for controllable interactive video worlds: let a world engine decide what changes, and let a video generator render how those changes appear.

\clearpage

\bibliographystyle{iclr2027_conference}
\bibliography{ref}

\clearpage
\appendix

\section{Video-to-Structure Annotation}
\label{app:annotation}
\subsection {Data Preparation}
This section describes how we collect, filter, and annotate the training videos.
\subsubsection {Collection and Filtering}
\textbf{Collection:} We collect videos from two sources:
\begin{itemize} 
\item \textbf{Online Datasets:} We rely on two large-scale datasets: VideoGen-1M~\citep{vidgen1m} and 4DNeX~\citep{4dnex}, which capture diverse objects, environments, and interactions.
\item \textbf{Gameplay Data:} We collect diverse gameplay videos from online sources, which provide structured object behaviors and interactions. 
\end{itemize}
\textbf {Filtering:} Once videos are collected, we apply a filtering process to ensure only high-quality, relevant videos are included in the dataset:
\begin{itemize}
\item \textbf{Resolution and Quality Filtering:} We only select videos with a minimum resolution of 720p, 16 FPS, and 81 frames.
\item \textbf{Content Relevance:} We keep only videos in which objects are identifiable and in motion.
\end{itemize}

\subsubsection{Data Annotation}
\begin{figure}[htbp]
    \centering
    \includegraphics[width=0.65\textwidth]{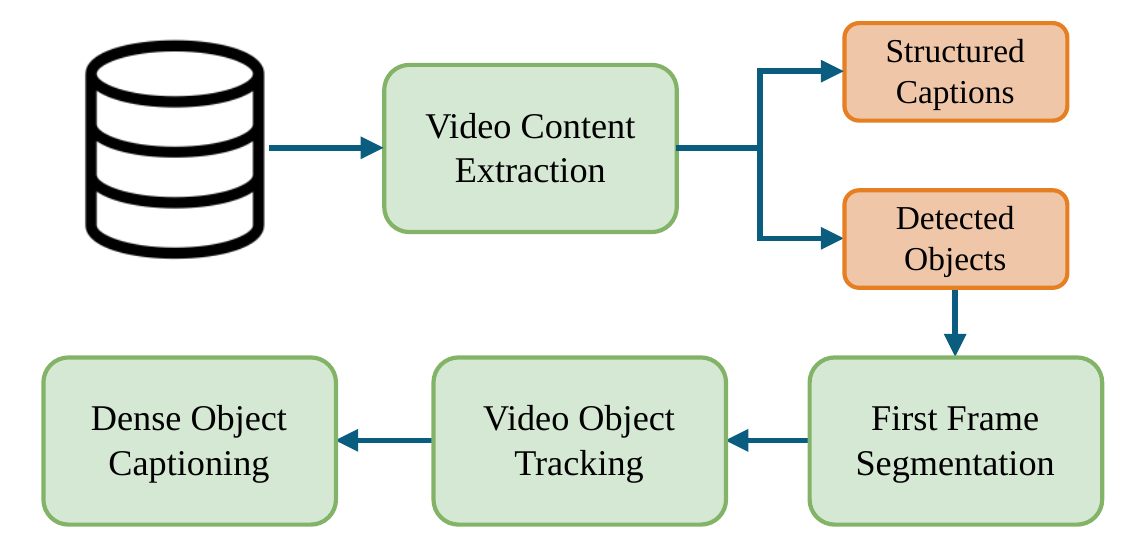}
    \caption{Video-to-Structure Annotation Pipeline.}
    \label{fig:fig6}
\end{figure}
\begin{figure*}[t]
\centering
\setlength{\tabcolsep}{2pt}
\renewcommand{\arraystretch}{1.0}

\parbox{0.98\textwidth}{
\scriptsize
\textbf{Global caption}
``The video features a person in a white t-shirt and brown pants attempting a skateboard trick. He is positioned on a concrete surface with a pink wall in their back. The individual starts by balancing on the skateboard, then jumps into the air while flipping the board. As they land, they lose balance and fall to the ground, ending up sitting on the pavement with the skateboard beside them.''
}

\vspace{0.5em}

\noindent
\makebox[\textwidth][c]{%
\includegraphics[width=0.19\textwidth]{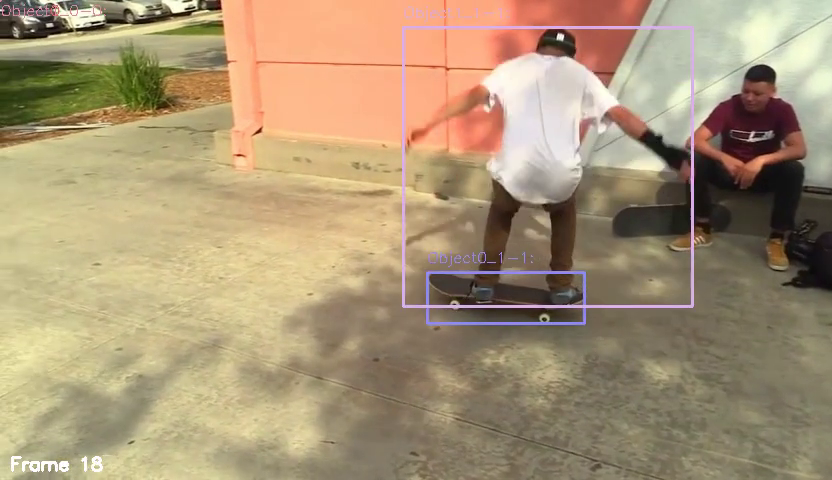}
\includegraphics[width=0.19\textwidth]{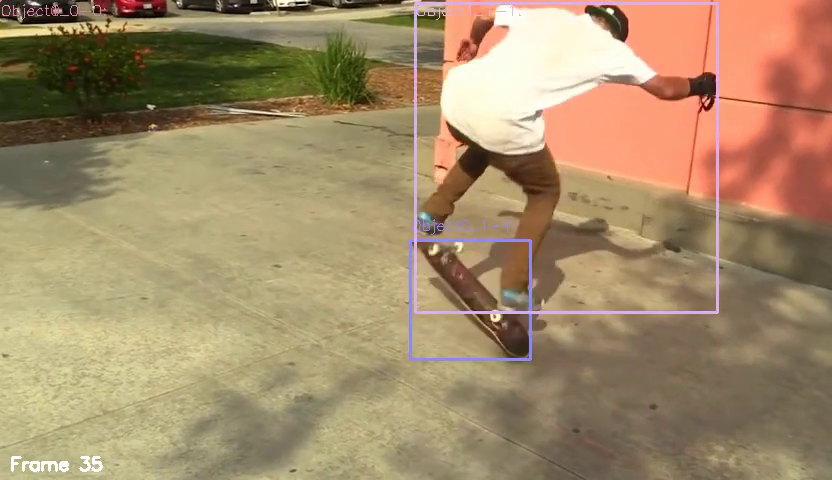}
\includegraphics[width=0.19\textwidth]{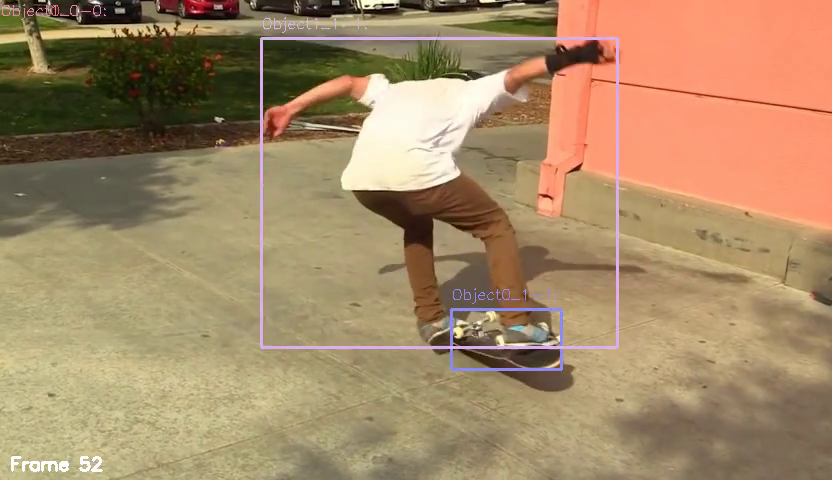}
\includegraphics[width=0.19\textwidth]{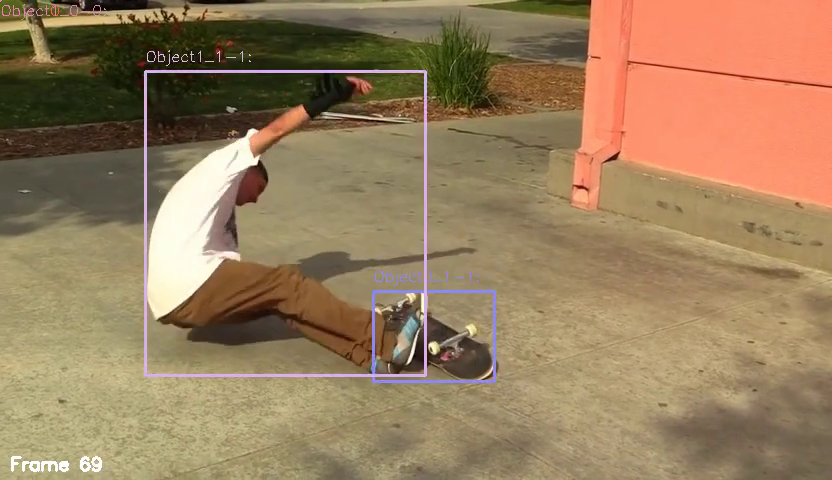}
\includegraphics[width=0.19\textwidth]{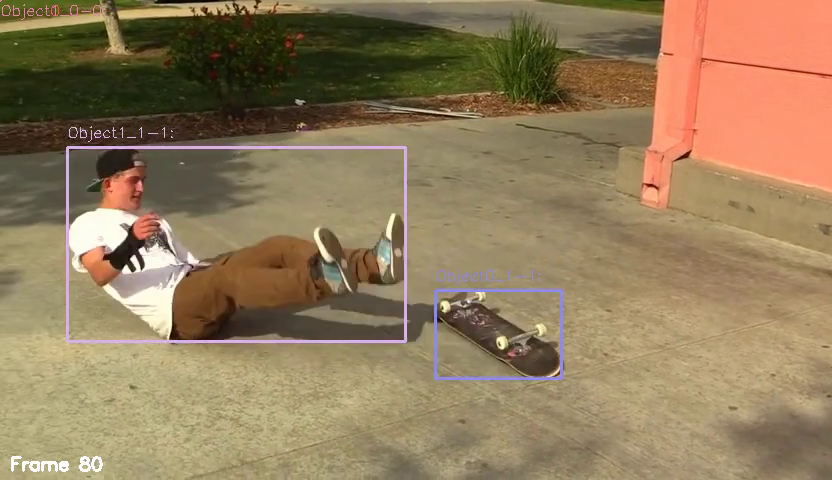}
}

\vspace{0.4em}

\parbox{0.98\textwidth}{
\scriptsize
\textbf{Object Descriptions:}

\textbf{Frame 18:}\\
\textcolor{blue}{\textit{object 1:}}
``The skateboard in the scene is a sleek, black deck with white wheels and trucks. The grip tape on top of the board appears to be worn from frequent use, suggesting it's well-loved by its owner. The underside of the board shows visible signs of wear as well, indicating that this isn't just any ordinary skateboard but one that has been used extensively for skateboarding stunts or tricks.'' \\
\textcolor{red}{\textit{object 2:}} ``The person in the scene is a young man skillfully riding his skateboard. He's wearing casual attire, including a white t-shirt and brown pants, which contrasts with the vibrant pink wall behind him. His black baseball cap adds to his urban style as he maneuvers through what appears to be an outdoor setting. The motion of skateboarding suggests energy and agility, capturing a dynamic moment that encapsulates youth culture and street sports.'' \\

\textbf{Frame 35:} \\
\textcolor{blue}{\textit{object 1:}}
``The skateboard in the scene is a classic design with four wheels, two at each end. It features white grip tape on its top surface and appears to be made of wood or composite material. The trucks are visible beneath the deck, connecting it to the wheels, while the underside shows the bottom layer where the bearings reside.'' \\ 
\textcolor{red}{\textit{object 2:}}
``The person in the scene is a young man performing an action on his skateboard. He's wearing casual attire, including a white t-shirt and brown pants, with black sneakers that match his cap. His posture suggests he might be executing a trick or maneuver as he moves along the sidewalk next to a pink wall.'' \\

\textbf{Frame 52:} \\
\textcolor{blue}{\textit{object 1:}}
``The skateboard in the scene is a classic design with four wheels, two at each end. It features a smooth deck and trucks that connect to the wheels for movement. The grip tape on top of the board provides traction when riding.'' \\
\textcolor{red}{\textit{object 2:}} ``The person in the scene is a young skateboarder wearing casual clothing, a white t-shirt paired with brown pants rolled up to the knees for ease of movement. The skateboarder has one arm extended behind for balance during the maneuver.'' \\

\textbf{Frame 69:} \\
\textcolor{blue}{\textit{object 1:}} ``The skateboard in the scene is a classic design with four wheels, two at each end. It features white trucks and yellow wheels that contrast against its black deck. The grip tape on top of the board appears to be worn out, suggesting it has been used extensively for skateboarding activities.'' \\
\textcolor{red}{\textit{object 2:}} ``The person in the scene is a young man losing balance on his skateboard after the trick. He is dressed casually, wearing a white t-shirt and brown pants paired with black sneakers, with his arms extended as he tries to stay upright.'' \\

\textbf{Frame 80:} \\
\textcolor{blue}{\textit{object 1:}} ``The skateboard in the scene is a classic design with four yellow wheels attached to its underside. The deck of the board features an intricate pattern, possibly graffiti or artwork, adding character and individuality to it. The trucks are silver-colored metal pieces that connect the wheels to the deck, providing stability for the rider's feet as they glide across surfaces.'' \\
\textcolor{red}{\textit{object 2:}} ``The person in the scene is a young man in casual attire, wearing a white t-shirt, brown pants, black sneakers, and a backward cap. He is sitting on the pavement after the fall, with the skateboard beside him.''

}

\vspace{0.3em}

\caption{
An example of our data annotation results from VideoGen-1M~\citep{vidgen1m}.
}
\label{fig:fig7}

\end{figure*}
\begin{figure*}[t]
\centering
\setlength{\tabcolsep}{2pt}
\renewcommand{\arraystretch}{1.0}

\parbox{0.98\textwidth}{
\scriptsize
\textbf{Global caption (Gameplay Data):}
``Main Object: The main object in the video is a racing car with blue and purple glowing accents, speeding through a grand, ornate track.; The car is moving forward on the track, accelerating and navigating through the course.;  Main Camera: The camera follows the car's motion, capturing the dynamic and fast-paced nature of the race.;  Landscape Environment: The background environment is a grand, ancient-style arena with golden pillars and a large circular emblem on the ceiling. The track is adorned with blue and pink glowing elements, and there are clouds in the sky.;  Time/Weather/Lighting: The video takes place in a time of day that appears to be either dawn or dusk, given the soft lighting and the presence of clouds in the sky. The weather conditions is clear with no visible rain or fog. The lighting is bright enough to clearly see the details of the race track and the surroundings, including the golden structures and the blue neon lights on the track.;  Targets: The secondary target in the video is a red car that appears on the right side of the track, slightly ahead of the main car.; The secondary target, a black car with blue accents and a lightning bolt symbol on the back, is moving in the same direction as the primary car. It maintains a consistent distance behind the primary car throughout the video.;  Interaction: The car navigates through the track, passing under a sign that reads "LAP 2" and continues to race towards the finish line.; The video features a racing game where the player's car is seen speeding through a grand, ornate track. The environment is richly detailed with classical architecture, including columns and statues, set against a bright, sunny sky. The track is adorned with various visual effects such as blue and pink glowing lines and circular markers indicating the path. The player's car navigates through the track, overtaking other opponents, and collects power-ups that enhance its speed and performance. The car's speed is indicated by a blue and purple exhaust effect, and the player maintains a high position in the race.''
}

\vspace{0.5em}

\noindent
\makebox[\textwidth][c]{%
\includegraphics[width=0.19\textwidth]{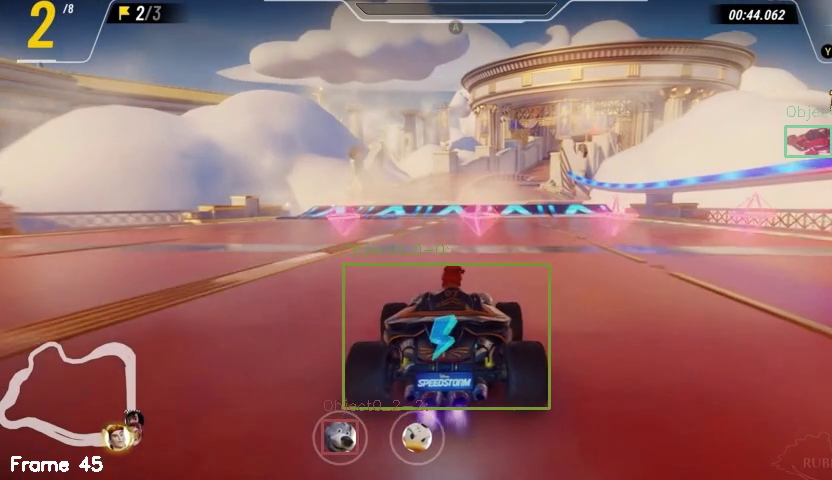}
\includegraphics[width=0.19\textwidth]{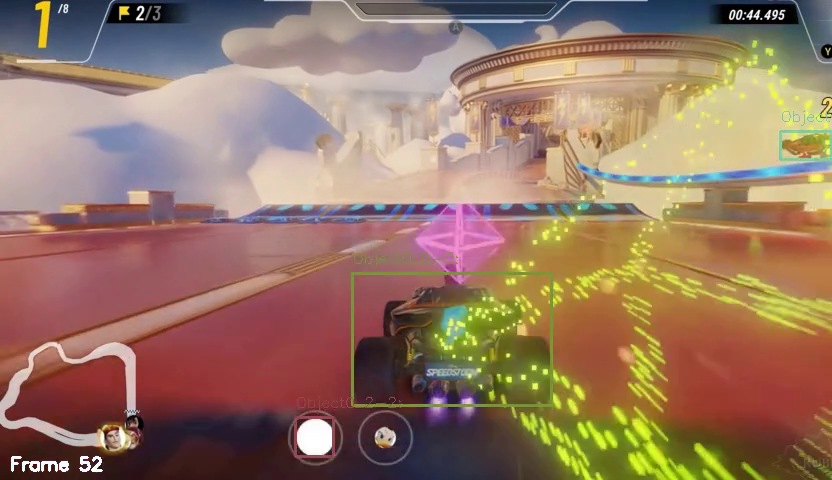}
\includegraphics[width=0.19\textwidth]{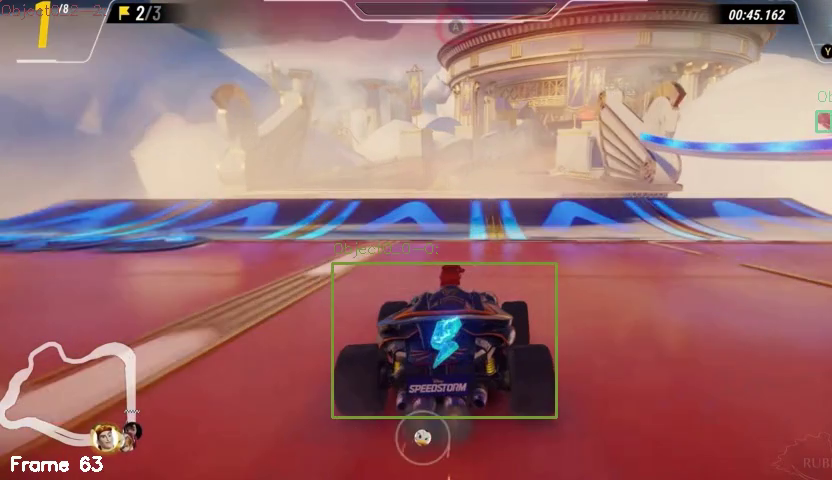}
\includegraphics[width=0.19\textwidth]{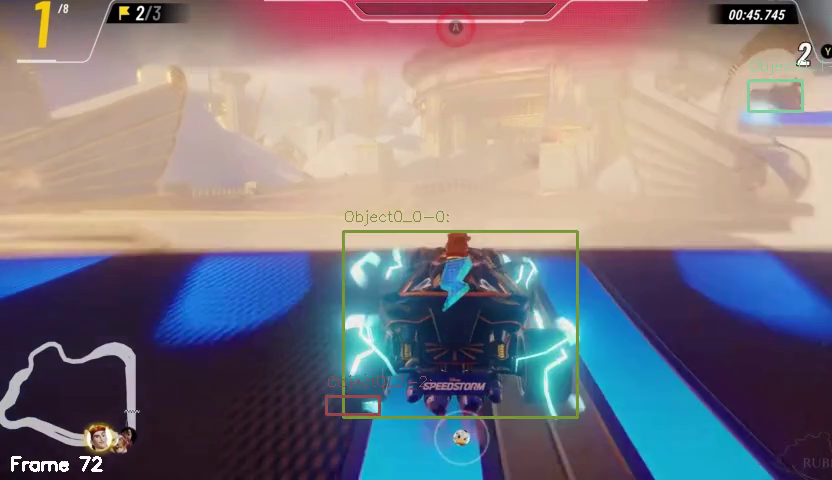}
\includegraphics[width=0.19\textwidth]{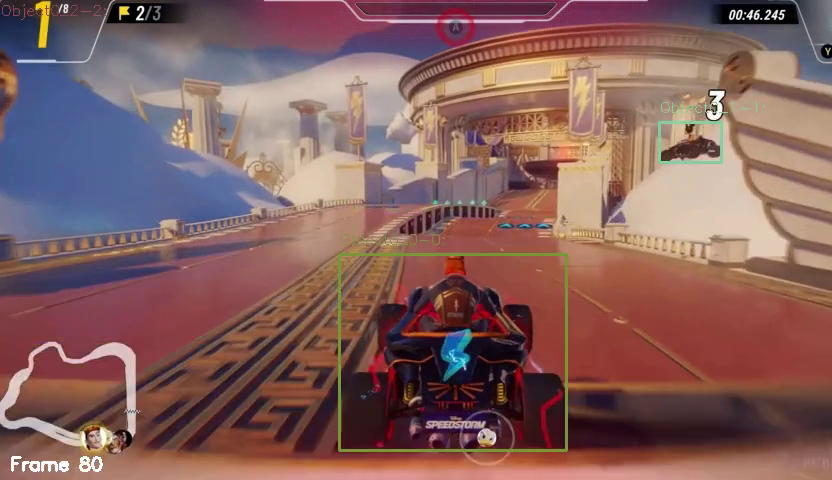}
}

\vspace{0.4em}

\parbox{0.98\textwidth}{
\scriptsize
\textbf{Object Descriptions:}

\textbf{Frame 45:}\\
\textcolor{blue}{\textit{object 1:}}
``The car in the scene is a high-performance vehicle, likely designed for speed and agility. It features sleek aerodynamic lines with vibrant colors that suggest it's built to stand out on the road or track. The rear of the car prominently displays an emblem resembling a lightning bolt, which could be indicative of its powerful engine capabilities. Additionally, there are visible exhaust pipes at the back, hinting at the performance-oriented nature of this machine.'' \\
\textcolor{red}{\textit{object 2:}} ``The car in the scene is a vibrant red, with its glossy finish reflecting light. The vehicle's design features sleek lines and curves that give it an aerodynamic appearance. It has tinted windows which add to its stylish look, while also providing privacy for passengers inside.'' \\

\textbf{Frame 52:} \\
\textcolor{blue}{\textit{object 1:}}
``The car in the scene is a futuristic vehicle with an aerodynamic design, featuring sleek lines and angular shapes. It has large black tires that contrast against its predominantly dark color scheme. The body of the car appears to be made from a metallic material, giving it a reflective quality. On top of the car, there's a glowing blue light source emanating bright particles or sparks around it, suggesting some form of energy emission or propulsion system.'' \\ 
\textcolor{red}{\textit{object 2:}}
``The car in the scene is a red sports car seen from behind, with a glossy finish and sleek, aerodynamic lines. It stays slightly ahead on the right side of the track.'' \\

\textbf{Frame 63:} \\
\textcolor{blue}{\textit{object 1:}}
``The car in the scene is a vibrant, high-performance vehicle with an aerodynamic design. It features striking blue and orange accents that give it a dynamic appearance. The hood of the car has a prominent lightning bolt graphic painted on it, adding to its aggressive look.'' \\
\textcolor{red}{\textit{object 2:}}
``The car in the scene is a striking blend of vibrant hues, with its body painted in a rich shade of red that transitions into a deep pink towards the front. The vehicle's design features sleek lines and curves, giving it an aerodynamic appearance. Its wheels are not visible from this angle, but one can imagine them to be well-maintained and complementing the overall aesthetic of the car.'' \\

\textbf{Frame 72:} \\
\textcolor{blue}{\textit{object 1:}} ``The car in the scene is a sleek, futuristic vehicle with an aerodynamic design. It features vibrant blue and black colors that give it a high-tech appearance. The body of the car has intricate details such as glowing lines running along its side, adding to its advanced look. On top of the car, there's a small structure or attachment resembling a lightning bolt, which stands out against the dark background.'' \\
\textcolor{red}{\textit{object 2:}} ``The car in the scene is a red sports car with a glossy finish. Its sharp angles and aerodynamic lines suggest speed as it continues along the track.'' \\

\textbf{Frame 80:} \\
\textcolor{blue}{\textit{object 1:}} ``The car in the scene is a sleek, high-performance vehicle with an aerodynamic design. It features vibrant colors and intricate detailing on its bodywork, including what appears to be glowing blue accents that add to its futuristic aesthetic. The rear of the car has prominent wings and exhaust pipes, suggesting it's built for speed and power.'' \\
\textcolor{red}{\textit{object 2:}} ``The car in the scene is a red sports car with a glossy body and rounded edges, still racing along the track ahead of the camera.''

}

\vspace{0.3em}

\caption{
An example of our data annotation results from gameplay videos.
}
\label{fig:fig8}

\end{figure*}

Our data labeling process follows a three-stage approach: \textbf{Video Content Extraction}, \textbf{Object Detection and Tracking}, and \textbf{Object Captioning}. For gameplay data, we implement an additional dedicated stage: \textbf{Structural Video Description with Element Breakdown}. Figures \ref{fig:fig7} and \ref{fig:fig8} show two representative examples of our annotation results, drawn from the online dataset and the gameplay dataset, respectively.
\newline
\newline
\textbf{Video Content Extraction.}
In the initial stage of labeling, we extract high-level content from each video using LLaVA-NeXT-Video~\citep{llava_next_video}. This model processes the entire video and performs two main tasks:
\begin{itemize}
    \item \textbf{Object Identification:} The model identifies the main objects present in each video to provide an initial list of visual entities appearing in the scene, such as humans, vehicles, animals, and inanimate objects.
\item \textbf{Full Video Description and Background Generation:} The model then generates a high-level description of the video, capturing not only the main objects but also the overall context of the scene. This includes background descriptions that characterize the environment, lighting, and any static elements in the scene that remain unchanged throughout the video.
\end{itemize}
\textbf{Object Detection and Tracking.}
Once objects have been identified by the video captioning model, we proceed to detect and track the position of these objects across video frames.
\begin{itemize}
\item \textbf{Initial Object Detection:} Grounding DINO~\citep{groundingdino} detects the identified objects in the first frame and produces their initial bounding boxes.
\item \textbf{Object Tracking:} After the initial detection, we use SAM~2~\citep{sam2} to track these objects across subsequent frames. SAM~2 allows us to maintain temporal consistency by tracking the same object across frames, even if it undergoes motion, partial occlusion, or appearance changes due to lighting or viewpoint shifts.
\end{itemize}
\textbf{Object Captioning.}
In the final labeling step, we crop each tracked object and caption it frame by frame with an image captioning model.
\begin{itemize}
    \item \textbf{Object Descriptions:} Each caption describes the object's identity and appearance (e.g., ``a red car''), which forms its persistent attribute.
\item \textbf{Temporal Captioning:} Each caption also describes the object's current status and action, with emphasis on status changes, which forms its dynamic status. For example, the red car might be described as ``a red car driving down the road'' in one frame and ``a red car stopping at a traffic light'' in another.
\end{itemize}
\textbf{Structural Video Description with Element Breakdown.}
For gameplay datasets, we enhance the standard video description by including additional elements that are specific to game environments and interactions. This breakdown captures both the visual aspects and the dynamic game mechanics. Specifically, we expand on the following key elements:
\begin{itemize}
    \item \textbf{Spatial Layout:} Descriptions of how objects are spatially arranged in the frame, such as their relative positioning and proximity to other objects or the camera. For example, ``The racing car is positioned in the center of the track, ahead of the player's car''.
\item \textbf{Action and Interaction Descriptions:} These provide detailed information about how objects interact with each other. We annotate the relationship between the primary player-controlled entity and the adversaries or objectives. This captures the behavioral logic of a game, where the player's actions directly influence the target's state and vice-versa. For example, ``The racing car aggressively closes the gap with the lead vehicle''.
\item \textbf{Background Descriptions:} Beyond individual object interactions, broader scene descriptions are generated, focusing on the environment or setting where the objects exist. For example, ``An urban street with tall skyscrapers, neon signs, and a crowd of pedestrians walking.''
\end{itemize}

\subsection{Example}
\label{app:annotation_examples}

Figures~\ref{fig:fig7} and~\ref{fig:fig8} illustrate how raw videos are converted into the object-centric state used by BoxWorld. Figure~\ref{fig:fig7} shows an open-domain clip from VideoGen-1M, where the annotation pipeline extracts a global video description, tracks salient objects across frames, and captions each object frame by frame. The captions describe both the persistent appearance of each object and its changing status, such as the skateboarder jumping, losing balance, and sitting on the pavement after the fall. Figure~\ref{fig:fig8} shows a gameplay clip, whose global description is further structured into the main object, camera, environment, targets, and interactions.

\section{Box World Engine Specification}
\label{app:box_world_engine}

A Box World Engine is an online state producer for the Object-Centric State Interface. It runs as a lightweight state machine that maintains entities, global variables, and authored update rules. At each step, the engine receives a user action, scripted policy output, or external control signal; evaluates the subset of rules that applies to the current state; updates object boxes, attributes, and dynamic statuses; and publishes the resulting state $\mathcal{S}_t$ to a state buffer. The renderer reads the latest state from the buffer asynchronously, while the engine continues stepping independently. This design keeps interaction logic editable and inspectable without coupling it to the renderer's denoising schedule or internal latent representation.

\paragraph{Engine components.}
Each engine instance is specified by four components. \textbf{Entities} define the active objects, including an identifier, semantic role, initial box or world-coordinate extent, persistent caption, and optional metadata. \textbf{Update rules} define how entities and optional global variables change in response to control signals and current state, including movement, activation, state transitions, visibility changes, object spawning, or other event-driven updates. \textbf{Camera and projection} parameters are used internally to map engine coordinates to image coordinates and to produce frame-level boxes. \textbf{State publishing} converts the updated engine state into the interface consumed by BoxWorld: a time-indexed stream of object entries $(\mathbf{b}_{t,i}, \mathbf{c}_i, \mathbf{d}_{t,i})$, where boxes are normalized frame coordinates, attributes describe persistent identity and appearance, and dynamic statuses describe the current status of each object.

\begin{algorithm}[t]
\caption{Generic Online Box World Engine}
\label{alg:box_world_engine}
\begin{algorithmic}[1]
\State Initialize entities $\mathcal{E}$, global state $\mathbf{g}$, and camera/projection parameters $C$
\State Initialize authored update rules $\mathcal{R}$ and output state buffer $\mathcal{B}$
\While{the interactive session is active}
    \State Receive user action, scripted policy output, or external control signal $u_t$
    \State Evaluate applicable rules $\mathcal{R}_t \subseteq \mathcal{R}$ from $(\mathcal{E}, \mathbf{g}, u_t)$
    \State Update entity boxes, attributes, dynamic statuses, and global state using $\mathcal{R}_t$
    \State Add, remove, activate, or deactivate entities if specified by the rules
    \State Project active entities to frame coordinates using $C$
    \State Publish $\mathcal{S}_t=\{(\mathbf{b}_{t,i}, \mathbf{c}_i, \mathbf{d}_{t,i})\}_{i=1}^{N_t}$ to $\mathcal{B}$
    \State The renderer asynchronously reads the latest state from $\mathcal{B}$ for video synthesis
\EndWhile
\end{algorithmic}
\end{algorithm}

\paragraph{Authoring modes.}
In our experiments, a coding agent generates each engine from user-specified rules and the shared engine template; the engine can also be provided directly by the user. An author specifies the entities, the control inputs they can receive, the rules that update object state, and the projection used to expose boxes in image coordinates. Changing the interaction behavior then amounts to editing the update rules or the labels attached to entities, while the renderer continues to consume the same Object-Centric State Interface.

\paragraph{Reuse and authoring cost.}
The interface and renderer are shared across worlds, while each engine supplies its own transition rules. An existing 2D or 3D engine can retain its native mechanics and physics, project its active entities into image-space boxes, and export the associated state descriptions. In our experiments, implementing a new rule system from the shared engine template requires 205--714 lines of code (median 351). A new visual or scene variant typically requires a 10--20-line JSON configuration, after which the engine produces the trajectories, event timing, and counterfactual branches programmatically.

\section{Additional Experimental Details and Results}
\label{app:additional_experiments}

\subsection{Implementation Details}
\label{app:implementation}
Stage 1 uses filtered open-domain videos from VideoGen-1M~\citep{vidgen1m} and 4DNeX~\citep{4dnex}, which provide diverse objects, environments, and mostly smooth object motion. Stage 2 uses 4,000 short gameplay clips collected online, where object roles, interactions, and event consequences are more explicit. We keep clips with sufficient resolution, frame rate, duration, visible moving objects, and reliable detection and tracking results, and discard clips with severe blur, unstable views, or ambiguous object identities (Appendix~\ref{app:annotation}).

We train the State Grounding Module for 20K iterations with batch size 32, using AdamW ($\beta_1=0.9$, $\beta_2=0.95$, weight decay $=0.01$) and linear warmup to a peak learning rate of $10^{-4}$. Stage 2 trains rank-16 LoRA adapters for 10K iterations with the same batch size and optimizer, using linear warmup to $3 \times 10^{-5}$ followed by cosine decay to 10\% of the peak learning rate. Stage 3 trains a separate rank-64 LoRA adapter for 400 DMD steps with all preceding components frozen. We use FlexAttention, bfloat16 mixed precision, and gradient checkpointing. At inference, the State Grounding Module is active only at noise levels above 0.8. The Box World Engines are created by Kimi K2.5~\citep{kimik25} from the shared engine template in Appendix~\ref{app:box_world_engine}.

\begin{figure}[p]
    \centering
    \includegraphics[width=\linewidth,height=0.9\textheight,keepaspectratio]{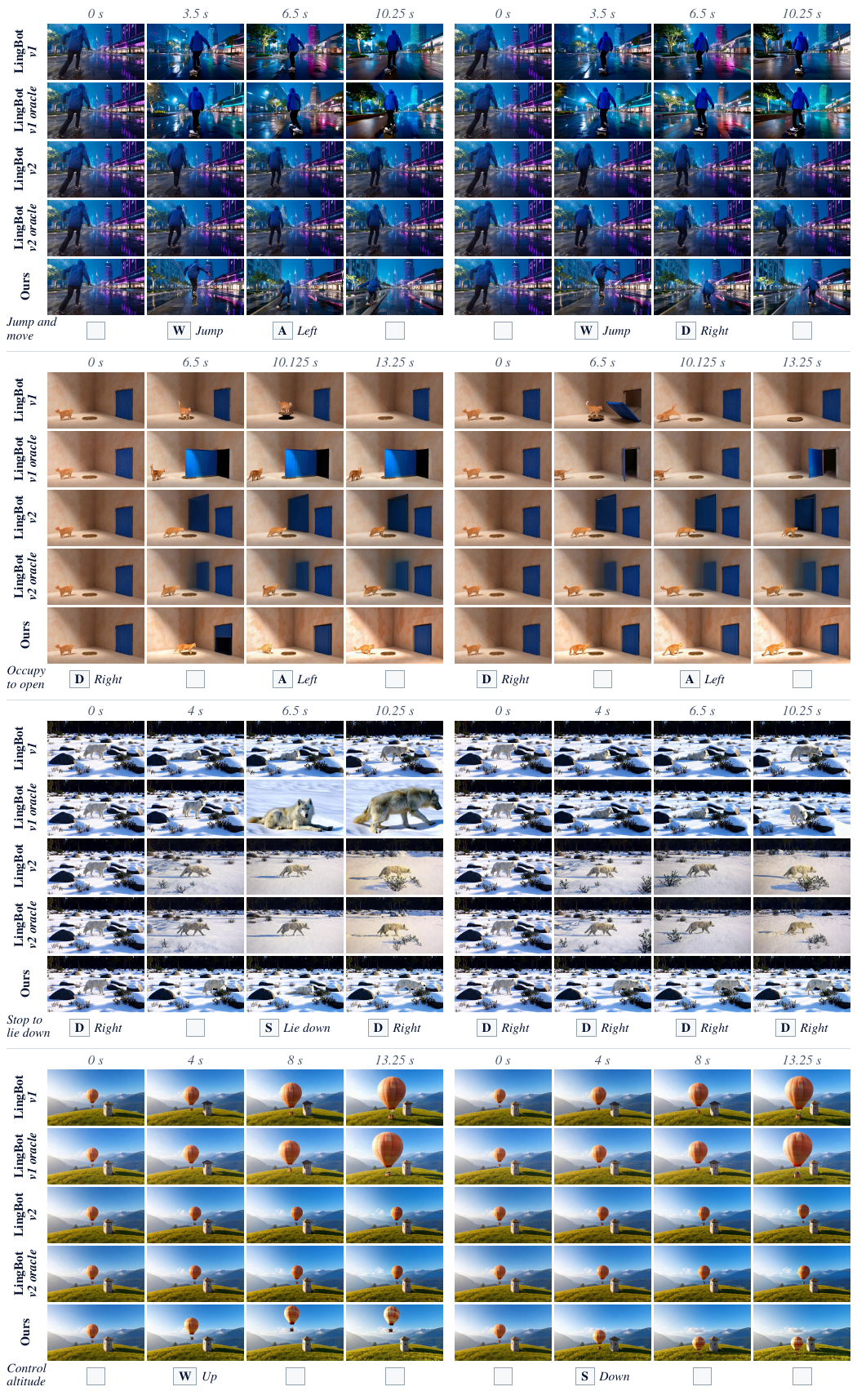}
    \caption{Additional qualitative comparisons on PairedStateBench with LingBot-v1, LingBot-v2, and their oracle variants: jumping and moving, occupying a pressure plate to open a door, stopping to lie down, and controlling altitude. Each case compares two branches with different user actions. Please refer to the \href{https://boxworld-web.github.io}{project page} for videos.}
    \label{fig:paired_comparison_appendix}
\end{figure}

\begin{figure}[p]
    \centering
    \includegraphics[width=\linewidth,height=0.9\textheight,keepaspectratio]{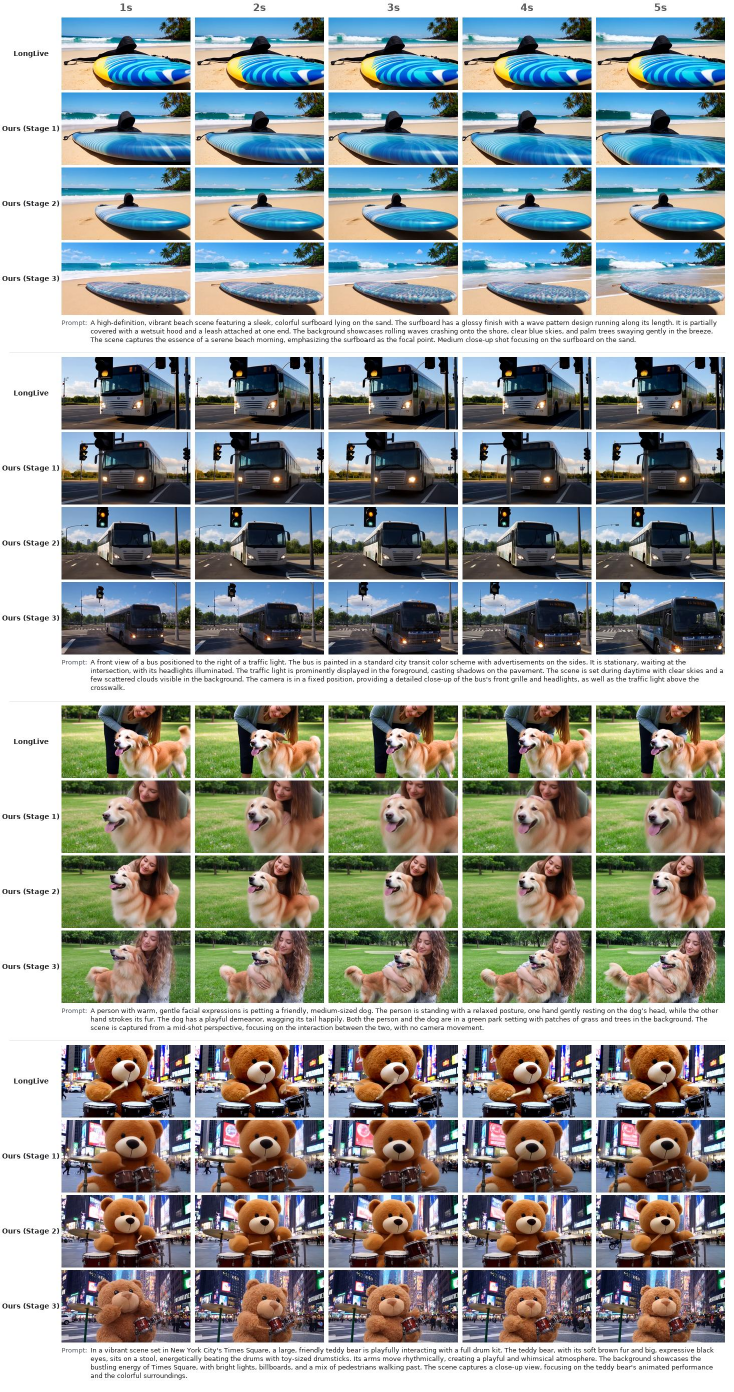}
    \caption{Additional qualitative VBench results comparing LongLive with the three stages of BoxWorld.}
    \label{fig:vbench_appendix}
\end{figure}

\subsection{PairedStateBench Protocol}
\label{app:authored_protocol}

\paragraph{Benchmark design.}
PairedStateBench tests whether a generated video realizes the consequences that user-defined rules assign to user actions. Each case pairs an engine-exported state trajectory with a specification of the events that must occur, the consequences that must not occur, and the state each object should hold over time. The two branches of a pair share the same scene and initial layout but receive different actions, so a model passes a pair only when it produces the intended outcome in both branches. The benchmark contains 87 cases from 16 scene families, such as object handling, credential-dependent doors, charging, gravity control, and pressure-plate doors, and includes 37 branch pairs. The easy, medium, and hard subsets contain 23, 28, and 36 cases, respectively.

\paragraph{Model inputs.}
BoxWorld uses the global prompt and the per-object box and dynamic-status sequence exported by the engine. LongLive uses the same global prompt without the object-level state sequence. LingBot-v1 uses the first frame of the case, the global prompt, and its native keyboard and camera inputs converted from the engine's navigation actions. LingBot-v2 uses the first frame, the global prompt, and camera conditions, as its public release exposes only camera-path conditioning. The oracle variants additionally receive a text description of the resolved state trajectory, which specifies the intended interaction consequences.

\paragraph{Metrics.}
\begin{itemize}
    \item \textit{Trigger}: the fraction of required events that appear in the video.
    \item \textit{Suppress}: the fraction of forbidden consequences that remain absent. Easy cases specify single actions without forbidden consequences, so Suppress does not apply to them.
    \item \textit{Paired-CF}: the fraction of pairs in which both branches realize their target outcomes.
    \item \textit{OnTime}: the fraction of required events whose located onset $f$ and engine boundary $f^\star$ satisfy
    \begin{equation}
        \left|\left\lfloor f/12\right\rfloor-
        \left\lfloor f^\star/12\right\rfloor\right|\leq 1,
    \end{equation}
    using decoded frame indices at 16 FPS.
    \item \textit{State-Fid}: whether each object holds its specified state within each state window. State-Fid excludes frames within the OnTime tolerance of each state boundary, so that it measures whether each state holds during its stable phase, while OnTime measures when the state switches.
    \item \textit{Task}: whether all action stages, forbidden-state constraints, event ordering, and final outcomes hold in the same video.
\end{itemize}
Single-video metrics are averaged within each video and then over videos.

\paragraph{Evaluation.}
Kimi K2.5~\citep{kimik25} and Qwen3.8-27B~\citep{qwen38} inspect chronological contact sheets of each decoded video and judge event outcomes, state agreement, and onset frames, and judgments on which the two models disagree are resolved by manual review. The two models disagree on 39 of the 522 evaluated videos (7.5\%), mainly where camera following or perspective scaling makes the direction or height of motion ambiguous. These judgments are compared with the engine reference to compute each metric. We will release the benchmark cases and evaluation code. Figure~\ref{fig:paired_comparison_appendix} shows additional qualitative comparisons.

\subsection{Video Quality}
\label{app:additional_video_quality}
\label{app:vbench_protocol}
\begin{table}[!htbp]
    \centering
    \footnotesize
    \setlength{\tabcolsep}{3.5pt}
    \caption{Video quality on VBench and 30-second VBench-Long. FPS measures generation throughput; baseline scores follow their reported evaluation settings.}
    \label{tab:video_quality}
    \begin{tabular}{lccccccc}
        \toprule
        & & \multicolumn{3}{c}{VBench} & \multicolumn{3}{c}{VBench-Long (30\,s)} \\
        \cmidrule(lr){3-5}\cmidrule(lr){6-8}
        Model & FPS$^{\uparrow}$ & Total$^{\uparrow}$ & Quality$^{\uparrow}$ & Semantic$^{\uparrow}$ & Total$^{\uparrow}$ & Quality$^{\uparrow}$ & Semantic$^{\uparrow}$ \\
        \midrule
        LTX-Video~\citep{ltx_video} & 8.98 & 80.00 & 82.30 & 70.79 & -- & -- & -- \\
        Wan2.1~\citep{wan} & 0.78 & 84.26 & 85.30 & 80.09 & -- & -- & -- \\
        SkyReels-V2~\citep{chen_skyreels-v2_2025} & 0.49 & 82.67 & 84.70 & 74.53 & 75.29 & 80.77 & 53.37 \\
        MAGI-1~\citep{sand_ai_magi-1_2025} & 0.19 & 79.18 & 82.04 & 67.74 & -- & -- & -- \\
        NOVA~\citep{deng_autoregressive_2024} & 0.88 & 80.12 & 80.39 & 79.05 & -- & -- & -- \\
        Pyramid Flow~\citep{jin_pyramidal_2024} & 6.7 & 81.72 & 84.74 & 69.62 & -- & -- & -- \\
        CausVid~\citep{causvid} & 17.0 & 81.20 & 84.05 & 69.80 & -- & -- & -- \\
        Self Forcing~\citep{self_forcing} & 17.0 & 84.31 & 85.07 & 81.28 & 81.59 & 83.82 & 72.70 \\
        LongLive~\citep{longlive} & 20.7 & 84.87 & 86.97 & 76.47 & 83.52 & 85.44 & 75.82 \\
        \midrule
        BoxWorld (Stage 1) & 20.7 & 83.30 & 84.46 & 78.62 & 81.72 & 82.81 & 77.36 \\
        BoxWorld (Stage 2) & 20.7 & 83.57 & 84.22 & 80.97 & 81.82 & 82.85 & 77.71 \\
        BoxWorld (Stage 3) & 20.7 & 83.76 & 84.82 & 79.54 & 83.00 & 84.34 & 77.67 \\
        \bottomrule
    \end{tabular}
\end{table}

We follow the standard VBench and VBench-Long protocols, generating videos at 832$\times$480 with four denoising steps. VBench videos have 81 frames, and VBench-Long videos extend to 30 seconds with the rolling KV cache of LongLive~\citep{longlive}. For each prompt, the object states are obtained by applying the annotation pipeline of Sec.~\ref{sec:video_to_structure} to a video generated by LongLive from the same prompt. Table~\ref{tab:video_quality} reports the VBench and VBench-Long results discussed in Sec.~\ref{sec:long_horizon}, together with additional non-real-time and real-time baselines. Compared with LongLive, Stage 1 improves the Semantic score from 76.47 to 78.62 while maintaining competitive Total and Quality scores, indicating that the State Grounding Module injects object-region conditions at high noise levels without substantially degrading visual quality. Stage 2 further improves the Semantic score to 80.97, and Stage 3 raises Total from 83.57 to 83.76 and Quality from 84.22 to 84.82. Figure~\ref{fig:vbench_appendix} shows qualitative examples comparing LongLive with the three stages.

\subsection{Layout-to-Video Generation}
\label{app:ytvis_protocol}
\begin{table}[!htbp]
    \centering
    \small
    \setlength{\tabcolsep}{4pt}
    \caption{Layout-to-video generation on YouTube-VIS 2021. Baseline scores are quoted from prior work, and all metrics follow the protocol of BlobGEN-Vid~\citep{blobgen_vid}.}
    \label{tab:youtubevis_grounding}
    \begin{tabular}{lcccccc}
        \toprule
        Method & Real-time & FVD$^{\downarrow}$ & mIOU$^{\uparrow}$ & rCLIP$_{t}^{\uparrow}$ & rCLIP$_{i}^{\uparrow}$ & rCFC$^{\uparrow}$ \\
        \midrule
        TrackDiffusion~\citep{trackdiffusion} & No & 464 & 0.492 & 0.273 & 0.804 & 0.940 \\
        LVD~\citep{lvd} & No & 558 & 0.281 & 0.261 & 0.803 & 0.861 \\
        VideoTetris~\citep{videotetris} & No & 590 & 0.166 & 0.267 & 0.799 & 0.423 \\
        BlobGEN-Vid~\citep{blobgen_vid} & No & 317 & 0.598 & 0.289 & 0.836 & 0.958 \\
        \midrule
        BoxWorld (Stage 3) & Yes & 62 & 0.572 & 0.291 & 0.837 & 0.965 \\
        \bottomrule
    \end{tabular}
\end{table}

We follow the YouTube-VIS 2021 setting and evaluation protocol of BlobGEN-Vid~\citep{blobgen_vid}, where ground-truth tracks provide layout boxes and per-track descriptions provide object attributes. Table~\ref{tab:youtubevis_grounding} reports the full comparison, and Table~\ref{tab:grounding_ablation} reports the grounding ablations under the same protocol.

\subsection{Additional Ablations}
\label{app:additional_ablations}
\begin{table}[!htbp]
    \centering
    \small
    \caption{Quantitative VBench ablations. The mask ablation is applied to Stage 3, and the adapter variants are compared with Stage 2; all variants use the four-step streaming sampler.}
    \label{tab:vbench_ablation}
    \begin{tabular}{lccc}
        \toprule
        Variant & Quality$^{\uparrow}$ & Semantic$^{\uparrow}$ & Total$^{\uparrow}$ \\
        \midrule
        Stage 3 & \textbf{84.82} & 79.54 & \textbf{83.76} \\
        \quad w/o spatial mask & 81.78 & 74.51 & 80.33 \\
        \midrule
        Stage 2 & 84.22 & \textbf{80.97} & 83.57 \\
        \quad w/o dynamics LoRA (Stage 1) & 84.46 & 78.62 & 83.30 \\
        \quad High-noise-only LoRA & 83.84 & 77.44 & 82.56 \\
        \quad Open-domain video LoRA & 82.68 & 79.16 & 81.98 \\
        \bottomrule
    \end{tabular}
\end{table}

\begin{table}[!htbp]
    \centering
    \small
    \setlength{\tabcolsep}{5pt}
    \caption{Grounding ablations on YouTube-VIS. Both ablations are applied to Stage 3; the attribute ablation replaces every per-box caption with ``object,'' retaining the global prompt, boxes, and original evaluation targets. All rows use four-step causal autoregressive generation.}
    \label{tab:grounding_ablation}
    \begin{tabular}{lccccc}
        \toprule
        Variant & FVD$^{\downarrow}$ & mIOU$^{\uparrow}$ & rCLIP$_t^{\uparrow}$ & rCLIP$_i^{\uparrow}$ & rCFC$^{\uparrow}$ \\
        \midrule
        Stage 3 & 62 & 0.572 & 0.291 & 0.837 & 0.965 \\
        \quad w/o spatial mask & 79 & 0.291 & 0.264 & 0.783 & 0.955 \\
        \quad w/o object attributes & 68 & 0.323 & 0.273 & 0.785 & 0.963 \\
        \bottomrule
    \end{tabular}
\end{table}

\begin{table}[!htbp]
    \centering
    \small
    \setlength{\tabcolsep}{4pt}
    \caption{Training stages on PairedStateBench. All variants use the same engine-exported state sequences.}
    \label{tab:stage_pairedstate}
    \begin{tabular}{lcccccc}
        \toprule
        Variant & Trigger$^{\uparrow}$ & Suppress$^{\uparrow}$ & Paired-CF$^{\uparrow}$ & State-Fid$^{\uparrow}$ & OnTime$^{\uparrow}$ & Task$^{\uparrow}$ \\
        \midrule
        Stage 1 & 87.3 & 60.3 & 64.9 & 83.4 & 83.8 & 49.4 \\
        Stage 2 & 82.0 & 85.7 & 67.6 & 83.6 & 81.5 & 64.4 \\
        Stage 3 & \textbf{92.2} & \textbf{90.5} & \textbf{97.3} & \textbf{91.6} & \textbf{91.2} & \textbf{83.9} \\
        \bottomrule
    \end{tabular}
\end{table}

\paragraph{Quantitative adaptation ablations.}
Table~\ref{tab:vbench_ablation} evaluates the adaptation settings on VBench. Removing the spatial mask from Stage 3 reduces Total from 83.76 to 80.33 and Semantic from 79.54 to 74.51. Among the adapter variants, the full Stage 2 model obtains the highest Semantic and Total scores, 80.97 and 83.57, while removing the Stage 2 adapters reduces Semantic from 80.97 to 78.62. Applying the adapters only at noise levels above 0.8 and training them on open-domain videos instead of gameplay videos yield Total scores of 82.56 and 81.98, respectively.

\paragraph{Training stages on PairedStateBench.}
Table~\ref{tab:stage_pairedstate} reports all PairedStateBench metrics for the three stages. Stage 1 already triggers most required events (87.3\% Trigger) but often produces forbidden consequences (60.3\% Suppress). Stage 2 raises Suppress to 85.7\% and Task to 64.4\%, while Trigger and OnTime decrease slightly to 82.0\% and 81.5\%. Stage 3 improves every metric, raising Paired-CF from 67.6\% to 97.3\% and Task to 83.9\%.

\paragraph{Spatial masks and object attributes.}
Table~\ref{tab:grounding_ablation} reports the full grounding ablations on YouTube-VIS. Both ablations are applied to Stage 3. Removing the spatial mask reduces mIOU from 0.572 to 0.291 and worsens FVD from 62 to 79. Replacing every object caption with the generic word ``object'' reduces mIOU to 0.323 and rCLIP$_t$ from 0.291 to 0.273, while rCFC remains similar (0.965 and 0.963).

\paragraph{Dynamic status.}
On PairedStateBench, replacing dynamic status with static descriptions lowers Trigger from 92.2\% to 42.5\% and OnTime from 91.2\% to 16.1\%. Without dynamic status, the renderer receives no signal of whether or when each object changes, so the target events are often missing or mistimed.

\section{Limitations}
\label{app:limitations}

By separating interaction reasoning from generative rendering, \texttt{BoxWorld} enables users to define object-level interaction rules and render their consequences in real time across diverse scenes and visual styles. Its current limitations mainly arise from the interface, the engine, and the training annotations. Although the Object-Centric State Interface, together with structured global prompts that describe the main objects, camera, environment, and interactions, covers most cases in our experiments, it may be insufficient for highly complex cases involving fine-grained geometry, articulated motion, or contact-rich interactions. Moreover, since the engine determines every interaction outcome, the rendered consequences are only as correct as the engine, and each new rule system requires its own engine implementation. The renderer is further trained on automatically annotated videos, where occasional annotation errors, such as objects missed by tracking, may lead to missing consequences or identity swaps. Future work can extend the interface with richer physical state and improve state annotation while keeping the interface lightweight and authorable.

\end{document}